\documentclass[acmlarge]{acmart}

\AtBeginDocument{%
  \providecommand\BibTeX{{%
    \normalfont B\kern-0.5em{\scshape i\kern-0.25em b}\kern-0.8em\TeX}}}

\copyrightyear{2022}
\acmYear{2022}
\setcopyright{acmlicensed}
\acmConference[UbiComp/ISWC '22 Adjunct]{Proceedings of the 2022 ACM International Joint Conference on Pervasive and Ubiquitous Computing}{September 11--15, 2022}{Cambridge, United Kingdom}
\acmBooktitle{Proceedings of the 2022 ACM International Joint Conference on Pervasive and Ubiquitous Computing (UbiComp/ISWC '22 Adjunct), September 11--15, 2022, Cambridge, United Kingdom}
\acmPrice{15.00}
\acmDOI{10.1145/3544793.3560391}
\acmISBN{978-1-4503-9423-9/22/09}

\usepackage{multirow}
\usepackage[shortlabels]{enumitem} 

\usepackage[hypcap=false]{caption}
\usepackage{subcaption}
\usepackage{amsmath}
\usepackage[capitalise]{cleveref}
\usepackage[font=small,skip=5pt]{caption}
\begin{document}

\title[Smart-Badge: A wearable badge with multi-modal sensors]{Smart-Badge: A wearable badge with multi-modal sensors for kitchen activity recognition}


\author{Mengxi Liu}
\orcid{0000-0003-0527-1208}
\author{Sungho Suh}
\orcid{0000-0003-3723-1980}
\author{Bo Zhou}
\orcid{0000-0002-8976-5960}
\author{Agnes Grünerbl}
\orcid{0000-0002-4156-7121}
\author{Paul Lukowicz}
\orcid{0000-0003-0320-6656}
\email{FirstName.LastName@dfki.de}
\affiliation{%
  \institution{German Research Center for Artificial Intelligence (DFKI)}
  \streetaddress{Trippstadter Str. 122}
  \city{Kaiserslautern}
  \country{Germany}
  \postcode{67663}}

\renewcommand{\shortauthors}{Liu and Suh et al.}

\begin{abstract}
Human health is closely associated with their daily behavior and environment. However, keeping a healthy lifestyle is still challenging for most people as it is difficult to recognize their living behaviors and identify their surrounding situations to take appropriate action. Human activity recognition is a promising approach to building a behavior model of users, by which users can get feedback about their habits and be encouraged to develop a healthier lifestyle. In this paper, we present a smart light wearable badge with six kinds of sensors, including an infrared array sensor MLX90640 offering privacy-preserving, low-cost, and non-invasive features, to recognize daily activities in a realistic unmodified kitchen environment. A multi-channel convolutional neural network (MC-CNN) based on data and feature fusion methods is applied to classify 14 human activities associated with potentially unhealthy habits. Meanwhile, we evaluate the impact of the infrared array sensor on the recognition accuracy of these activities. We demonstrate the performance of the proposed work to detect the 14 activities performed by ten volunteers with an average accuracy of 92.44 \% and an F1 score of 88.27 \%. 

\end{abstract}




\maketitle

\section{Introduction}

Human activity and the surrounding situation in daily life have a substantial effect on human health and life quality. Although keeping a healthy lifestyle has emerged as a popular topic in the crowd, it is still difficult for many people, especially old people, to recognize their behavior and identify situations around them daily. Human activity recognition has been a promising tool for both users understanding their behavior and doctors diagnosing potential diseases, by which life quality of humans could be increased significantly. Therefore, human activity recognition is an important research direction in pervasive computing \cite{xu2019innohar} and has been widely investigated over the decades. With the rapid development of sensor technology and artificial intelligence algorithms, various solutions for human activity recognition based on novel sensor modality \cite{oguntala2019smartwall,jiang2020novel,cheng2010active} and machine learning algorithms \cite{matsuyama2021deep,ordonez2016deep,gunthermann2022slow} have been proposed to extract comprehensive context from human activities predominantly, including body position-related, body action-related and body status-related context \cite{bian2022state} with the use of wearable devices or ambient sensors, by which user's lifestyle can be evaluated, and a behavior model can be build-up, which will make significant sense for detection of anomalies possibly relevant to well-being \cite{kehler2018systematic}, meanwhile, the feedback from the behavior model can, in turn, encourage users to develop a healthy lifestyle.  

Indoors is an environment where people spend much time; human activities in an indoor environment can reflect their living habits directly. Thus, indoor activity recognition has been widely used in many intelligent systems, from smart homes and smart health to smart security \cite{zhu2018indoor}. For example, human activities in the kitchen are closely associated with their dietary habits. In most cases, the frequency of open refrigerators and microwave ovens can indicate food intake frequency in the long term. Eating too much, too frequently, or abnormal eating time could form an unhealthy dietary habit, which can lead to many diseases like diabetes \cite{volkow2011reward}, obesity \cite{mayer1967regulation}, and cardiovascular disease \cite{bazzano2002fruit}. Kitchen scene context-based activity recognition thus is a promising approach for diet monitoring, and dietary treatment, also helpful for developing smart kitchens as a part of smart home \cite{luo2019kitchen}. In addition, it also provides meaningful information for people to understand their dietary habits, which play an essential role in promoting a healthy lifestyle through interventions \cite{kamachi2021prediction}. 

However, compared to human activity recognition in other application scenarios, kitchen activity recognition is still not widely explored. In this paper, we introduce a smart badge integrated with multi-modal sensors to recognize human activity in a kitchen scenario. We designed the multi-sensor-based hardware platform to be packaged in a light badge, and thus easily attachable to the user's chest, and includes six different sensors and two microcontrollers. It is worth mentioning that we use an infrared array sensor (thermal sensor) instead of a camera to avoid privacy issues. In addition, we adopt a multi-channel convolutional neural network (MC-CNN) \cite{yang2015deep} based on data and feature fusion methods and evaluate the performance of different sensors for activity recognition in the kitchen.

The contributions of this work are summarized as follows:
\begin{itemize}
\item A hardware platform of smart badge with 6 sensors for 14 kinds of common human activity recognition in kitchen was proposed.  
\item Two kinds of multi-channel convolutional neural network for data fusion and feature fusion are adopted and provide high recognition accuracy of 14 kinds of human activity.
\item The performances of different sensor in kitchen activity recognition are investigated. 
\end{itemize}

The remaining of this paper is organized as follows. \cref{sec:relatedwork} presents the related works in the fields of kitchen activity recognition and multi-modal sensor platform. The detailed presentation of hardware implementation is introduced in \cref{sec:hardware}. \cref{sec:eval} presents the collected data and the quantitative experimental results. Finally, \cref{sec:discussion} shows the analysis of the experimental results and \cref{sec:conclusion} addresses conclusions and future works.

\section{related work}
\label{sec:relatedwork}
\subsection{Kitchen activity recognition}
Although kitchen activity recognition is not so widely explored as other application scenarios, there are still some solutions for activity recognition in a kitchen scenario proposed over the decades as kitchen activity of humans closely related to their dietary habits. The approaches can be grouped into vision-based and sensor-based. Vision-based methods combined with machine learning are widely utilized for kitchen activity recognition. For instance, Bansal et al. \cite{bansal2013kitchen} used a dynamic SVM-HMM hybrid model to predict nine cooking activities from video information with a recognition accuracy of 72 \%. Lei et al. \cite{lei2012fine} proposed a study for fine-grained recognition of kitchen activities with the use of RGB-D (Kinect-style) cameras, and the proposed system can robustly track and accurately recognize detailed steps through cooking activities. Although the vision-based method has showed a remarkable performance for HAR in different application scenarios \cite{natarajan2008online,yang2012recognizing,du2015hierarchical,yang2016super}, it often requires high computation capability and a well-lighted environment. Besides, privacy issues also prevent its widespread use. With the thriving development of sensor technology and pervasive computing, sensor-based HAR with privacy well protected is becomming more and more popular \cite{wang2019deep}.
Luo et al. \cite{luo2019kitchen} demonstrated a minimal and non-intrusive, low-power, low-cost radar-based sensing network system recognizing 15 kinds of activities with an accuracy of 92.8 \%. 
However, this solution lacks flexibility as the sensing network system should be deployed in the kitchen and can only detect the activity in a limited space. The wearable device has shown more flexible advantages over such distributed sensor system. 
Yasser et al. \cite{mohammad2017dataset} presented a dataset for 15 kinds of kitchen activity recognition using smartwatch accelerometers. Besides, they achieved a classification precise of 97.6 \% with the use of CNN based approach, which shows that a weareable device has a great potential in kitchen activity recognition. 

\subsection{Multi-modal sensors platform in human activity recognition}
Compared to the limitation of computer vision technology in the human activity recognition area, like space-time limitations, easy invasion of user privacy, and high energy consumption, the sensor-based methods with many advantages like compact, low cost, and high computational power have become the focus of attention \cite{qiu2022multi}. 
A single special-purpose sensor can only recognize single series activities. It also suffers from low robustness in most cases because most sensors have limitations due to sensor deprivation, limited spatial coverage, occlusion, imprecision, and uncertainty \cite{gravina2017multi}. Besides, an unhealthy lifestyle is usually the result of much bad behavior. Thus, using a single sensor for human activity recognition is not a perfect option in many scenarios. The concurrent use of multiple sensors for human activity recognition provides a practical solution for complex activity recognition, and improvement of recognition accuracy \cite{aguileta2019multi}.
For example, Zhang et al. \cite{zhang2020necksense} designed a necklace with multiple embedded sensors, such as a proximity sensor, an ambient light sensor, and an inertial measurement Unit (IMU) sensor. The necklace can detect the eating activity more accurately after augmenting the proximity sensor data with the ambient light and IMU sensor data.
Bharti et al. \cite{bharti2018human} proposed a HuMAnsystem with five kinds of sensors such as IMU, temperature, air pressure, and humidity sensor, as well as Bluetooth beacon, which are deployed on different parts of the body. This system showed that 21 complex activities at home could be detected with high accuracy.
Gravina et al. \cite{gravina2019emotion} demonstrated a system based on body-worn inertial sensors combined with a pressure sensor to monitor in-seat activities, by which four ordinary basic emotion-relevant activities were recognized with high accuracy. 
These studies about human activity recognition shows that complex human activity can benefit from multi-modal sensors information, which can increase system reliability and improve recognition accuracy. In this paper, we design a wearable smart badge base on multiple modal sensor platform with six different sensors to recognize kitchen activities, which could help users know and understand their habits.



\section{Hardware implementation }
\label{sec:hardware}

\begin{figure}
     \centering
     \begin{subfigure}[b]{0.49\textwidth}
         \centering
         \includegraphics[width=\textwidth]{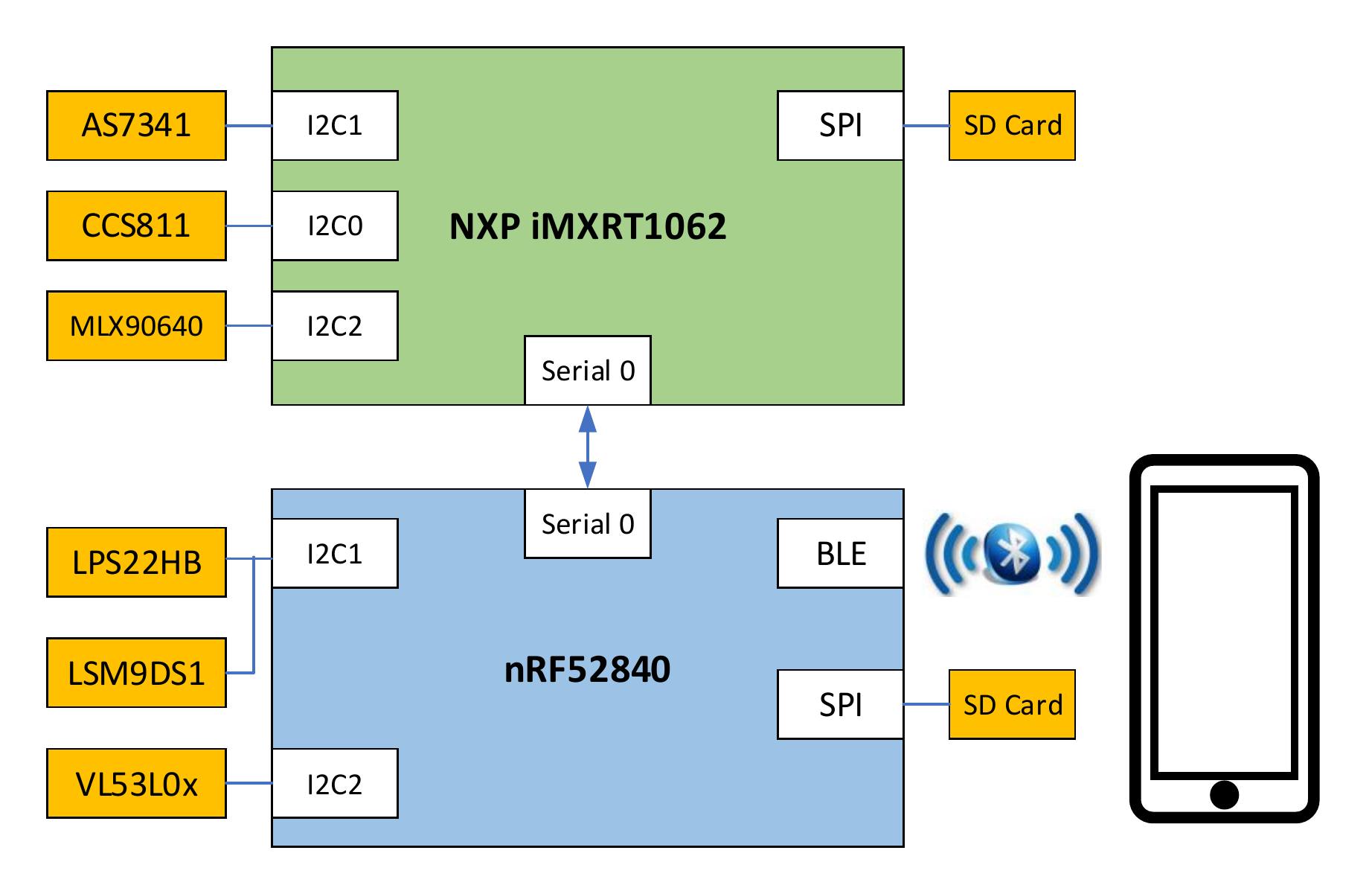}
         \caption{Hardware architecture}
         \label{fig:hwdesign}
     \end{subfigure}
     \hfill
     \begin{subfigure}[b]{0.49\textwidth}
         \centering
         \includegraphics[width=\textwidth]{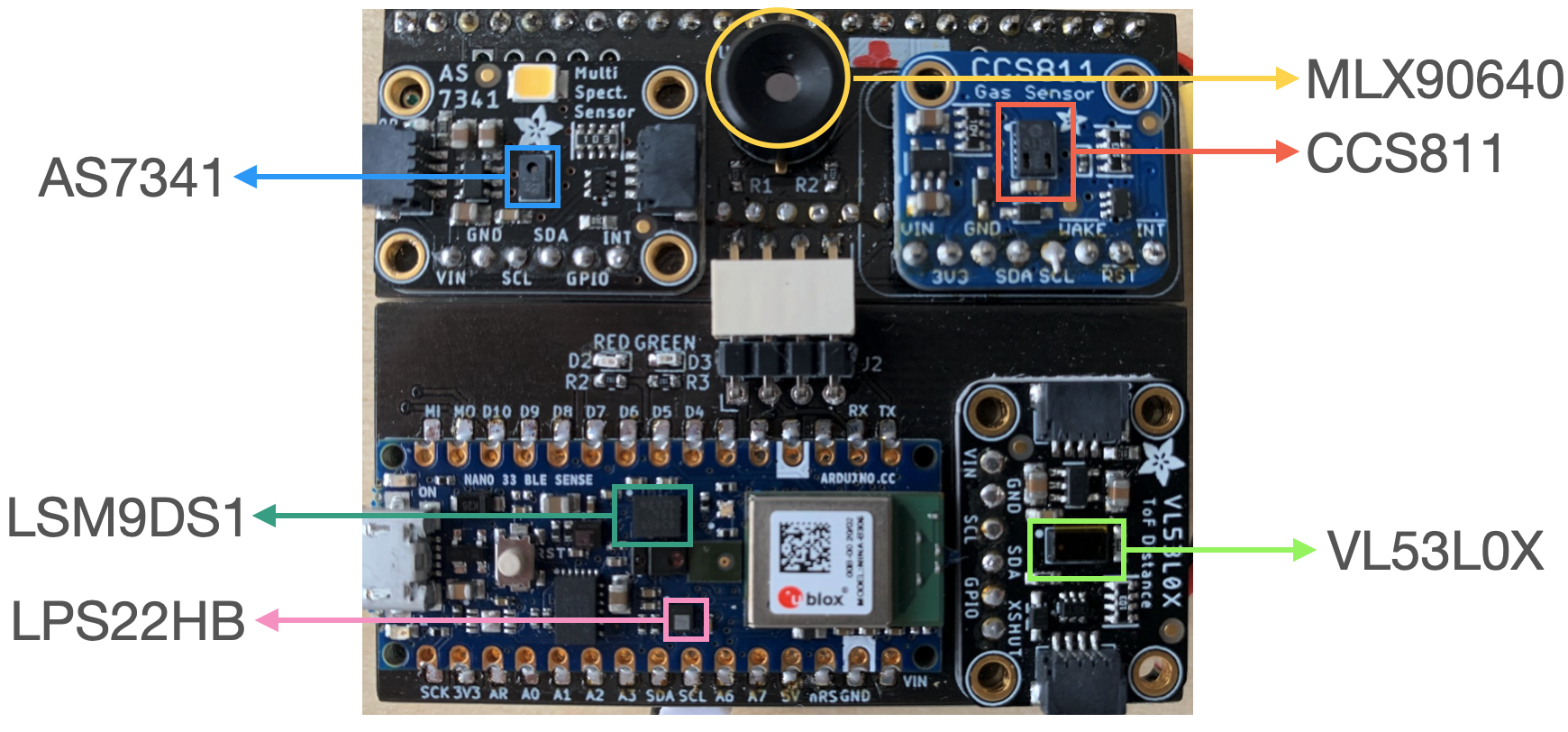}
         \caption{PCB Prototype}
         \label{fig:pcb}
     \end{subfigure}
        \caption{Hardware design of smart badge (\textbf{AS7341}: optical sensor (10 channel features). \textbf{CCS811}: digital gas sensor (2 channel feature). \textbf{MLX90640}: 32x24 pixels thermal IR array (768 channel feature). \textbf{LPS22HB}: air pressure sensor (1 channel feature). \textbf{LSM9DS1}: 9-axis iNEMO inertial module (9 channel feature). \textbf{VL53L0X}: Time-of-Flight ranging sensor (1 channel feature))}
        \label{fig:systemdesign}
\end{figure}

The ergonomics and ease of use are of paramount importance for a wearable device, which can not be a burden to the user during wearing \cite{bianchi2019iot}. Therefore, a smart light wearable badge was designed in this work, which can be attached to many parts of the body flexibly and easily, like shoulders, hips, and chest. As shown in \cref{fig:systemdesign}, the smart badge hardware system consists of four main components such as sensor module, microcontroller module, data logger module as well as data transmission module. Six different sensors (IMU, optical sensor, gas sensor, air pressure sensor, thermal IR array, and Time of Flight (ToF) ranging sensor) are connected to two microcontrollers via the I2C interface. 791 channel data from these six sensors, including body motion and ambient information, are sampled. 
Since the sample rate of these sensors varies considerably, for instance, the sample rate of IMU sensor LSM9D01 is up to 400 Hz in fast mode, while the sample rate of gas sensor CCS811 is only 4 Hz. If all sensors are connected to one I2C bus system, the sampling rate will be decreased dramatically. Furthermore, the number of communication interfaces of one micro-controller is limited for connecting the six sensors separately. Two microcontrollers (NXP iMXRT1062 based on high-performance ARM Cortex-M7 processor core and nRF52840 integrated with Bluetooth 5 on Arduino nano 33 board) are utilized to acquire data. The NXP iMXRT1062 microcontroller can operate at speed up to 600 Mhz, which provides a high-performance platform to run the tiny machine learning in future work. Since higher sampling rates can provide more precise information, while it also results in more power consumption, on a trade-off performance and power consumption, a low sampling rate is selected.
We use the NXP iMXRT1062 microcontroller to read data from thermal IR array sensor MLX90640, gas sensor CCS811, and optical sensor AS7341 with a speed of 3 Hz. The nRF52840 microcontroller reads data from the rest sensors at 12 Hz. After synchronization and interpolation, the final sampling rate of all sensors is 6 Hz. The sensor data can be transmitted to other devices by Bluetooth or stored locally on SD cards. Besides, configuration information like the activity label and time synchronization can be sent from a smartphone to microcontrollers via Bluetooth. The data exchange between two microcontrollers is via serial port. \cref{fig:pcb} shows the PCB prototype of the hardware platform with an entire dimension of 56$\times$64 mm. The working current of the whole system is 154 mA. Besides, to improve the user-friendly experience, a light 3D printed package of this hardware system is designed as shown in \cref{fig:prototype}. The velcro is attached to the back side so that the smart badge can be wearied on many body parts easily. 

\section{Evaluation}
\label{sec:eval}
\subsection{Data collection}
\begin{figure}
     \centering
         \begin{subfigure}[b]{0.4\textwidth}
         \centering
         \includegraphics[width=\textwidth]{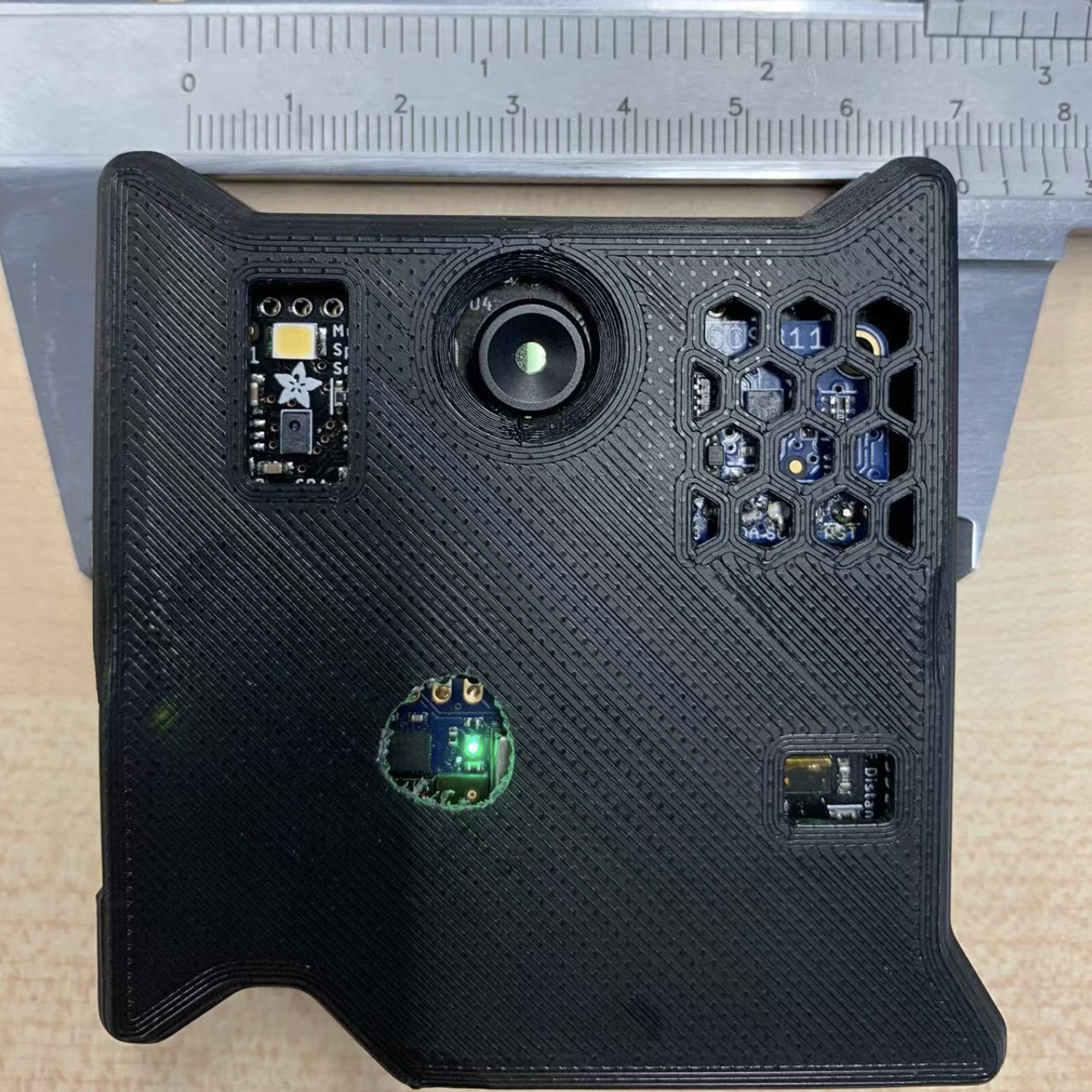}
         \caption{Smart badge prototype}
         \label{fig:prototype}
     \end{subfigure}
     \hfill
     \begin{subfigure}[b]{0.45\textwidth}
         \centering
         \includegraphics[width=\textwidth]{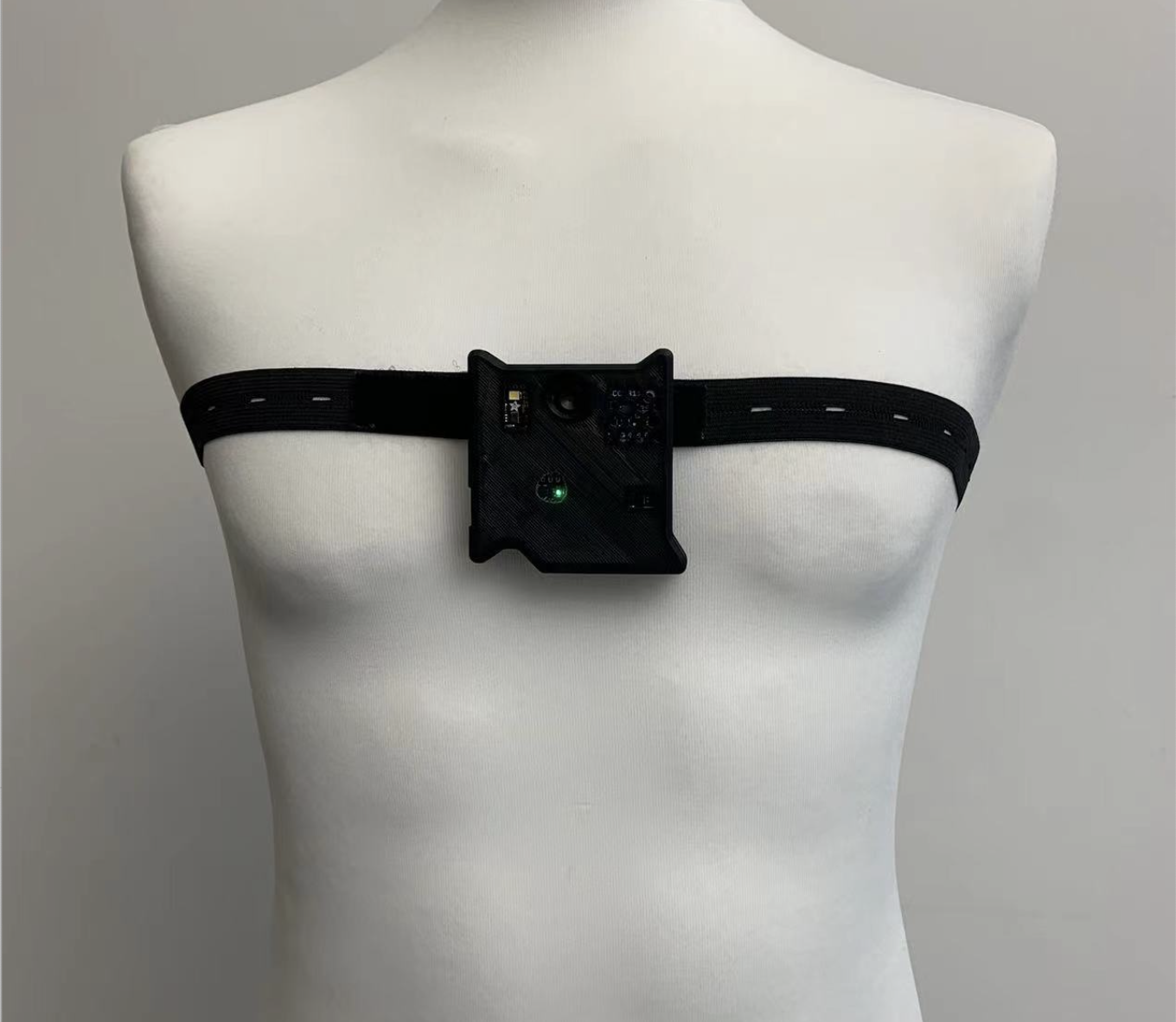}
         \caption{Wearing style}
         \label{fig:wearing}
     \end{subfigure}
        \caption{Smart badge prototype and wearing style}
        \label{fig:smartbagde}
\end{figure}

\begin{figure}
    \centering
    \includegraphics[width=\textwidth]{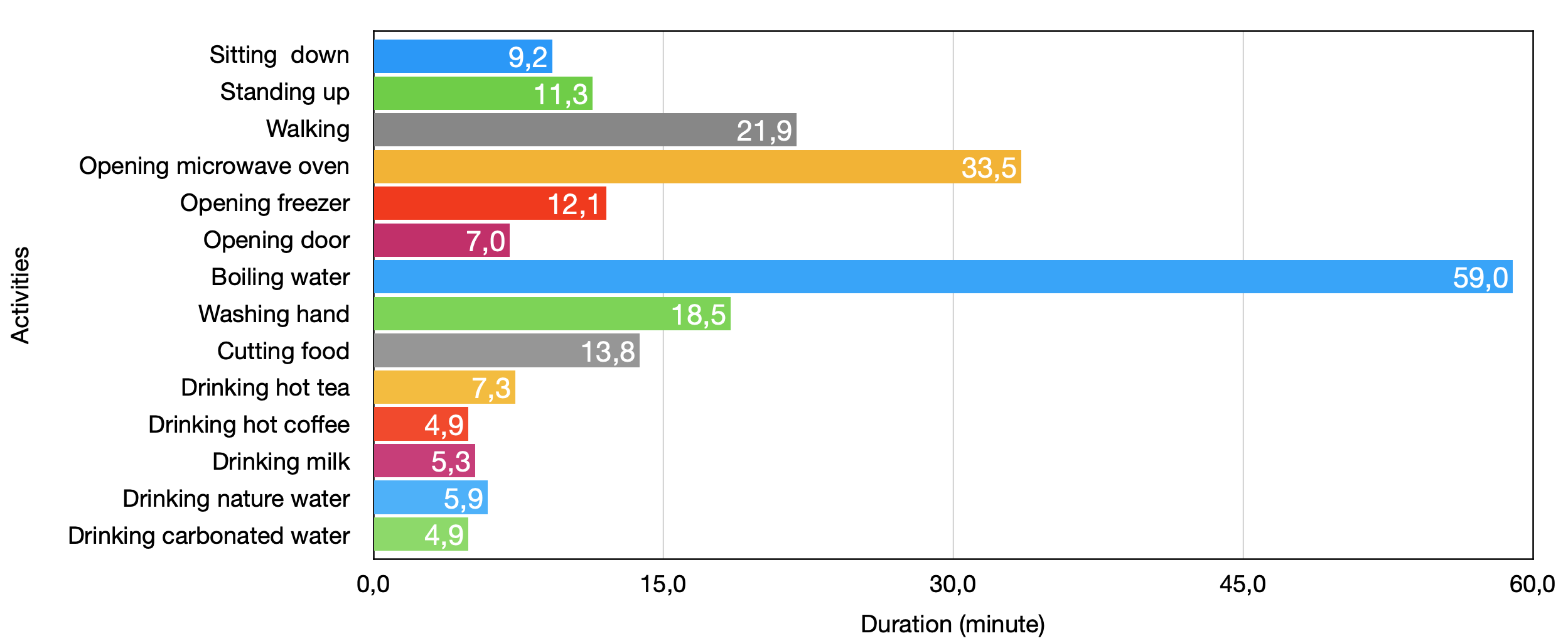}
    \caption{Dataset information}
    \label{fig:dataset}
     
\end{figure}

To evaluate the feasibility of our proposed approach for human activity recognition in the kitchen, we asked ten volunteers (seven male and three female) to put the smart badge on their chest as shown in \cref{fig:wearing} and perform fourteen daily activities as described in \cref{tab:activity} in a realistic unmodified kitchen environment. Most activities to recognition in this experiment strongly correlate with users' dietary habits. For example, food preparation is often related to opening a freezer, opening a microwave, cutting food, or boiling water. It should be possible to infer the user's eating time and frequency by recognizing those activities. Fluid intake activities like drinking different beverages are of utmost importance for the nutrition balance of the body. Thus, five typical kinds of drinking activities were recorded in this experiment.
All drinks were poured into clear glasses of typical size and shape. The experiment was divided into five sessions, and the whole sessions lasted around one hour per volunteer. Meanwhile, the label of each activity was transmitted from the smartphone to the micro-controller via Bluetooth interface by the experiment conductor. The sensor and label data were stored together on the local SD card. After collecting the data from ten volunteers, the 14 activities were segmented according to the labels. The null class was removed. \cref{fig:dataset} shows the whole duration time of each activity performed by ten volunteers. Each activity takes a different time to finish. For example, boiling water often takes several minutes, while drinking water only costs several seconds normally. Therefore, the entire duration time of each activity in this dataset is imbalanced.

\begin{figure}
     \centering
     \begin{subfigure}[b]{0.24\textwidth}
         \centering
         \includegraphics[width=\textwidth]{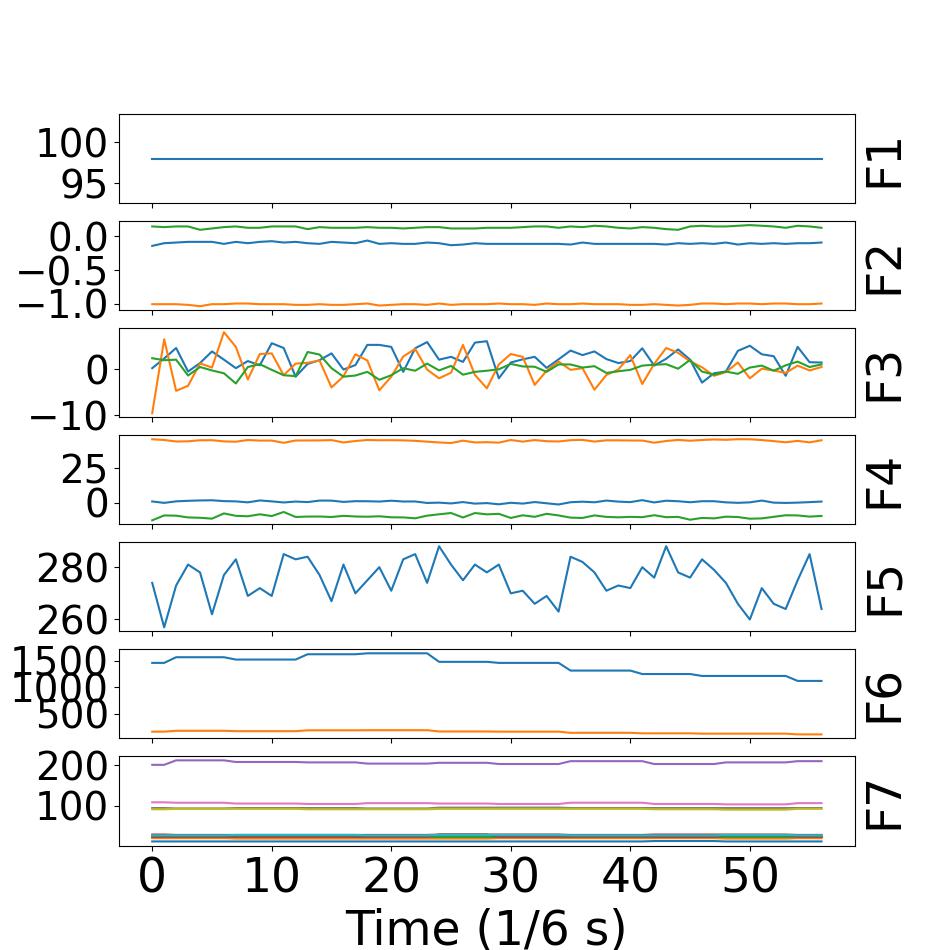}
         \caption{Sitting down}
         \label{fig:label0}
     \end{subfigure}
     \hfill
     \begin{subfigure}[b]{0.24\textwidth}
         \centering
         \includegraphics[width=\textwidth]{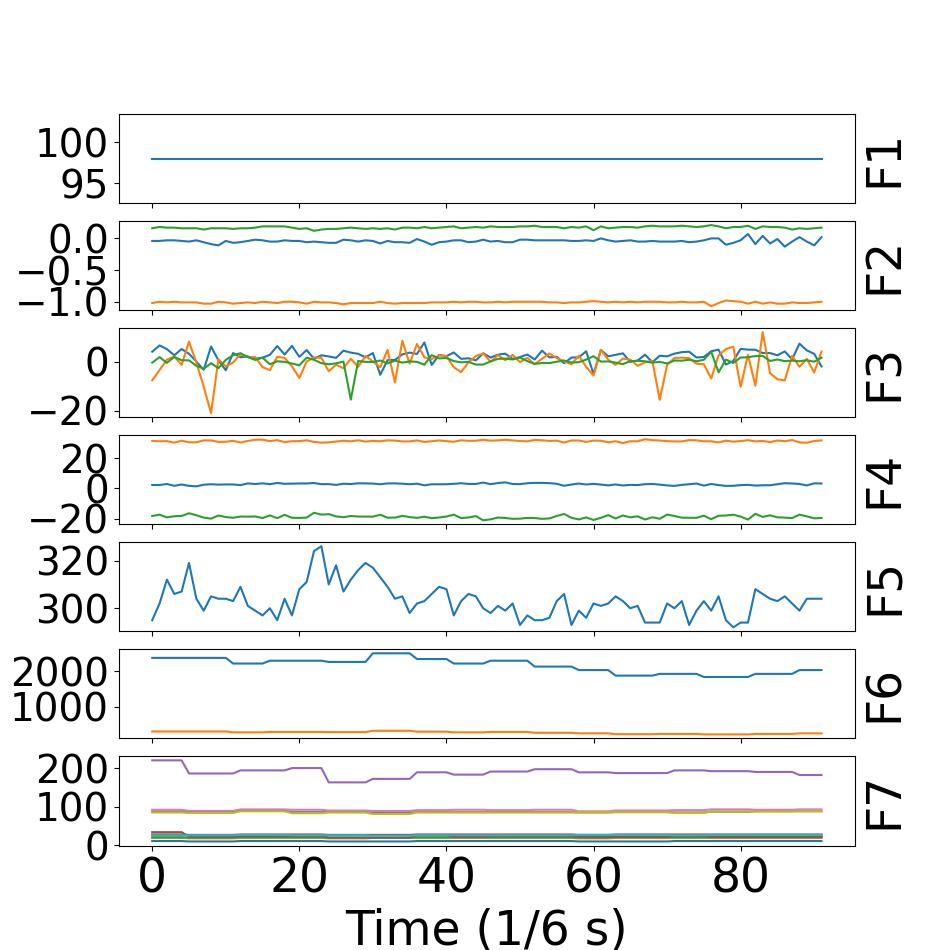}
         \caption{Standing up}
         \label{fig:label1}
     \end{subfigure}
     \hfill
     \begin{subfigure}[b]{0.24\textwidth}
         \centering
         \includegraphics[width=\textwidth]{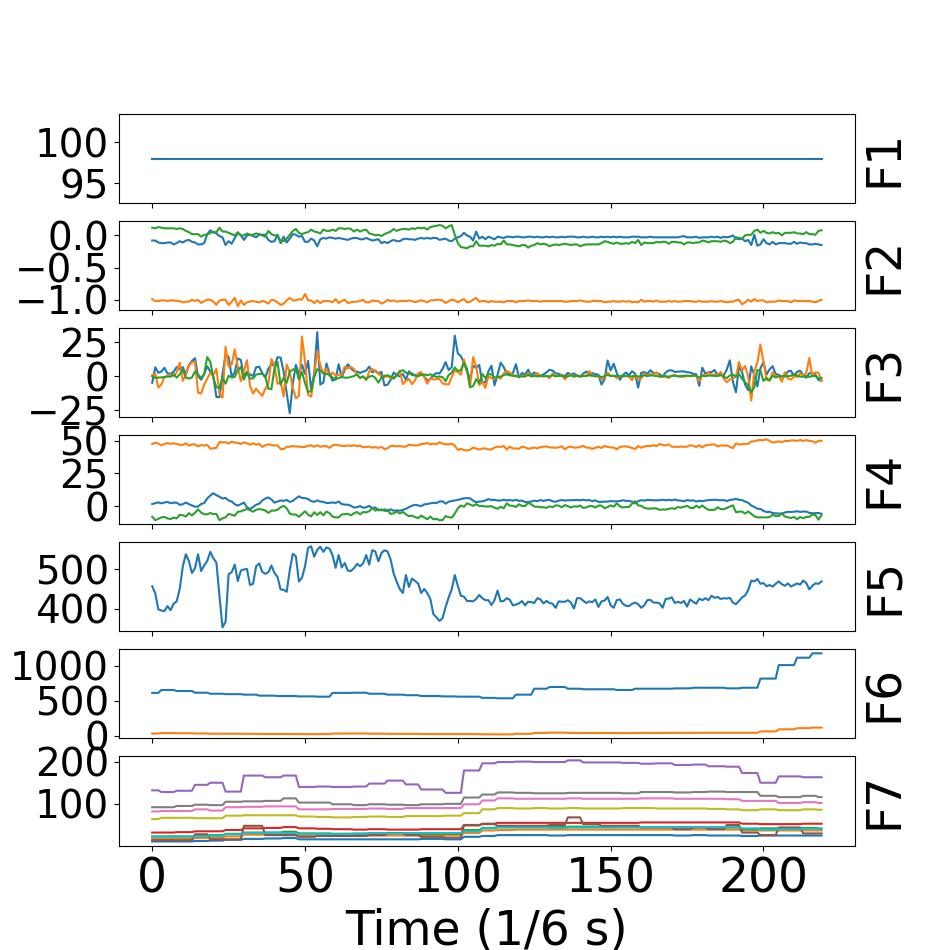}
         \caption{Walking}
         \label{fig:label2}
     \end{subfigure}
     \hfill
     \begin{subfigure}[b]{0.24\textwidth}
         \centering
         \includegraphics[width=\textwidth]{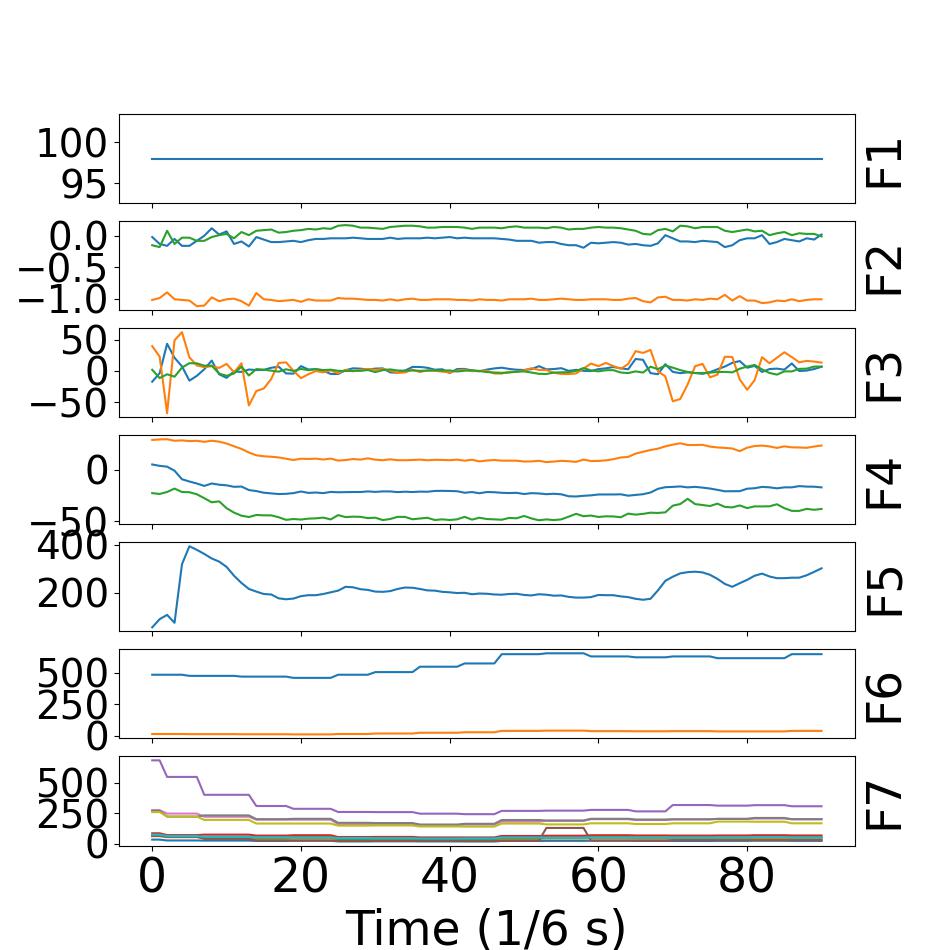}
         \caption{Opening microwave}
         \label{fig:label3}
     \end{subfigure}
     
     \hfill
     \begin{subfigure}[b]{0.24\textwidth}
         \centering
         \includegraphics[width=\textwidth]{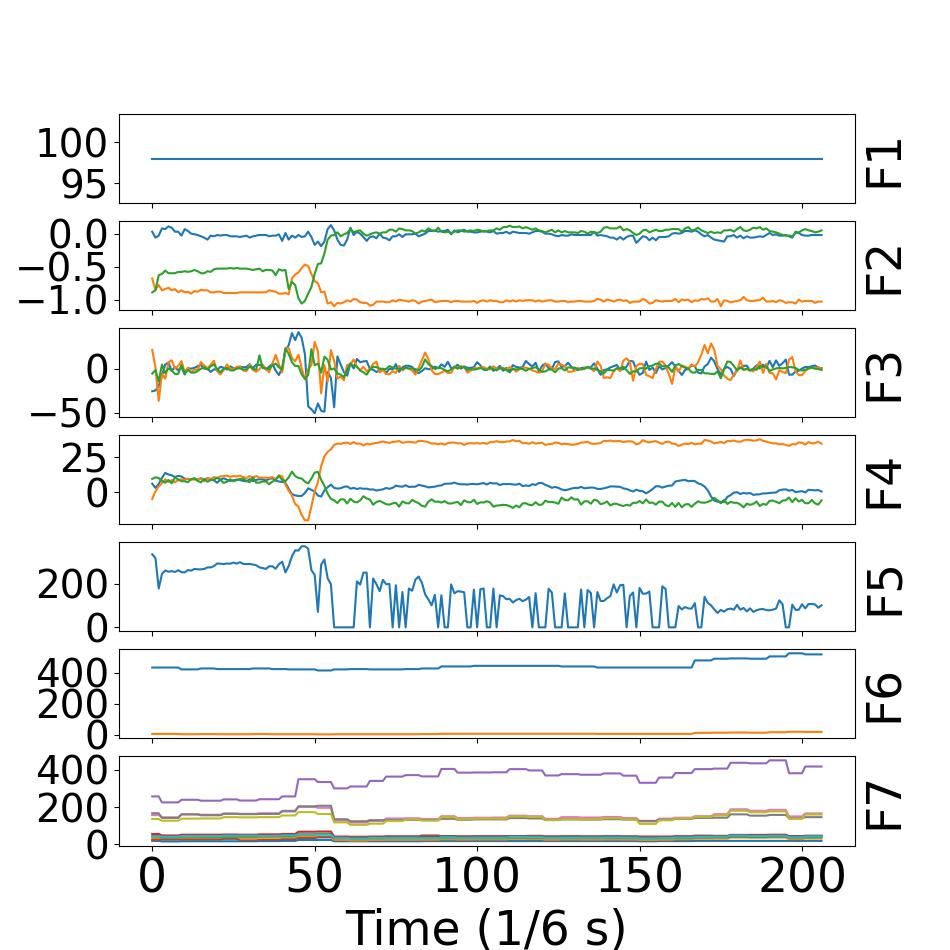}
         \caption{Opening freezer}
         \label{fig:label4}
     \end{subfigure}
     \hfill
     \begin{subfigure}[b]{0.24\textwidth}
         \centering
         \includegraphics[width=\textwidth]{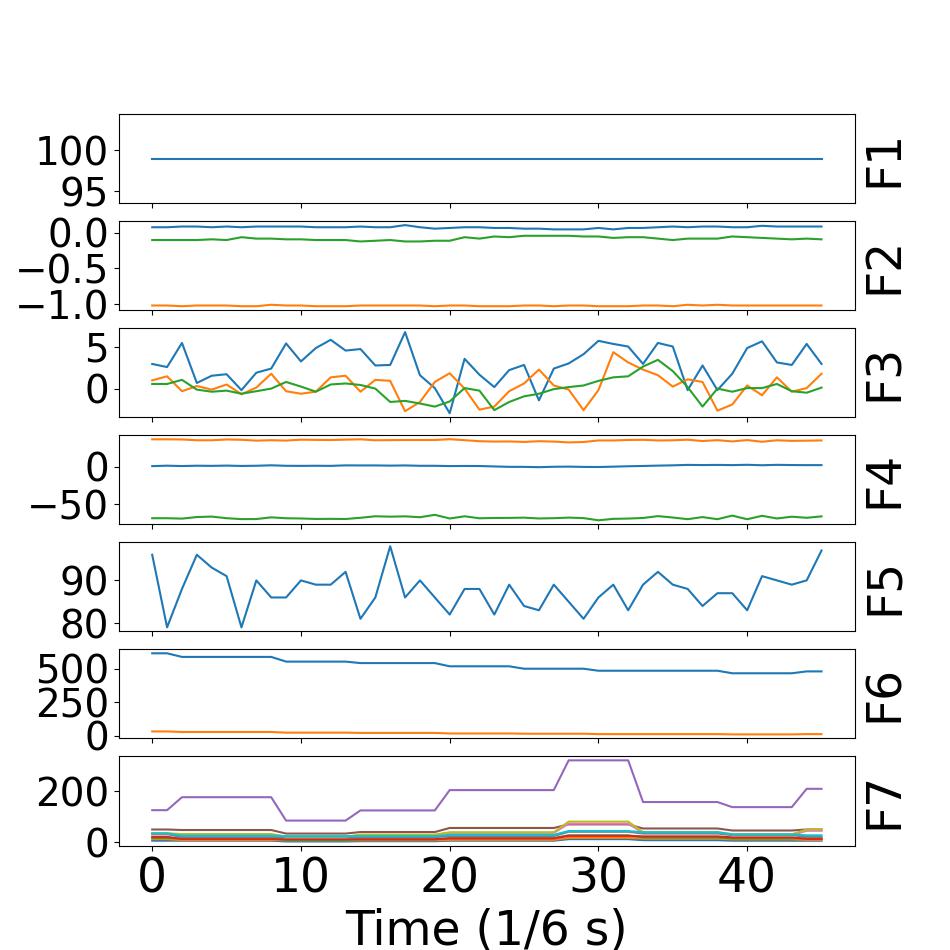}
         \caption{Opening door}
         \label{fig:label5}
     \end{subfigure}
     \hfill
     \begin{subfigure}[b]{0.24\textwidth}
         \centering
         \includegraphics[width=\textwidth]{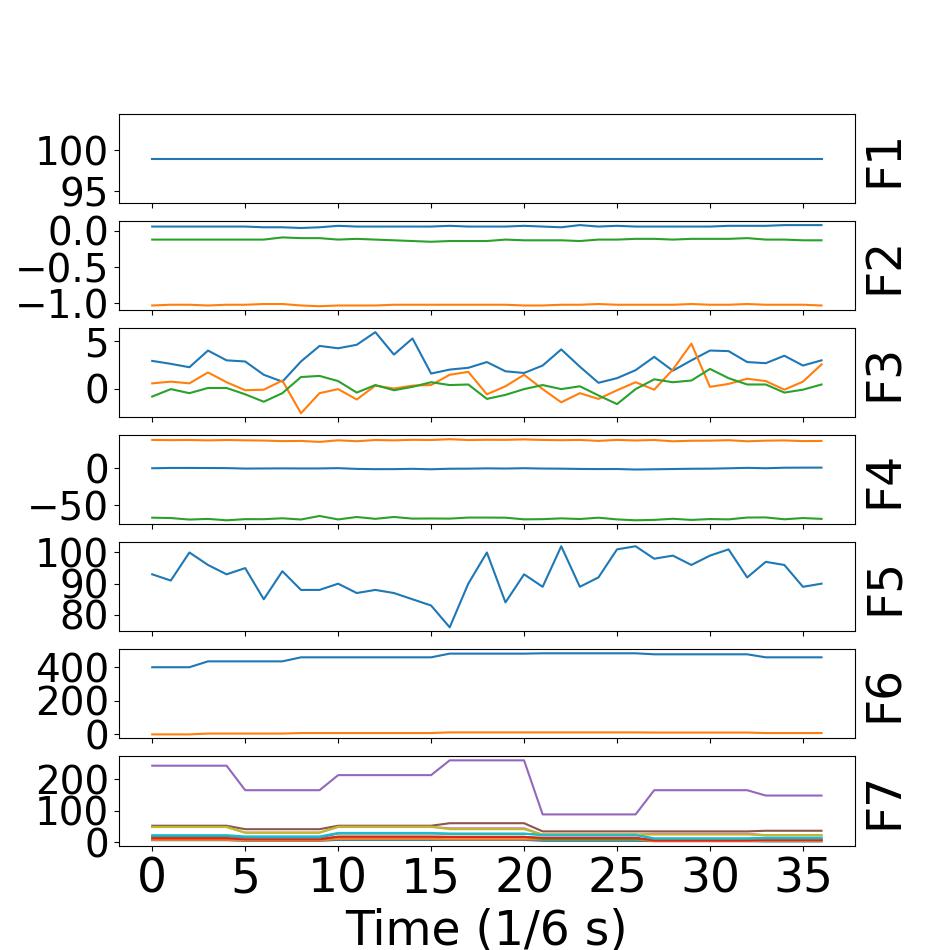}
         \caption{Boiling water}
         \label{fig:label6}
     \end{subfigure}
     \hfill
     \begin{subfigure}[b]{0.24\textwidth}
         \centering
         \includegraphics[width=\textwidth]{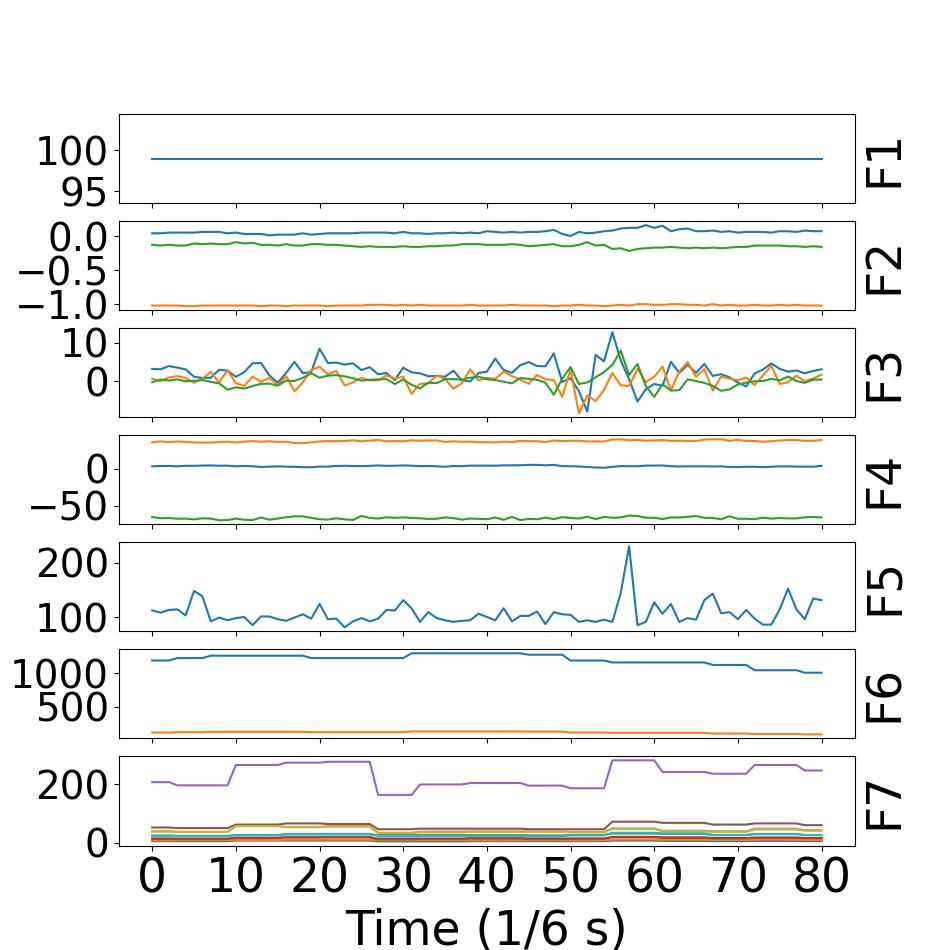}
         \caption{Washing hand}
         \label{fig:label7}
     \end{subfigure}
     
     \hfill
     \begin{subfigure}[b]{0.24\textwidth}
         \centering
         \includegraphics[width=\textwidth]{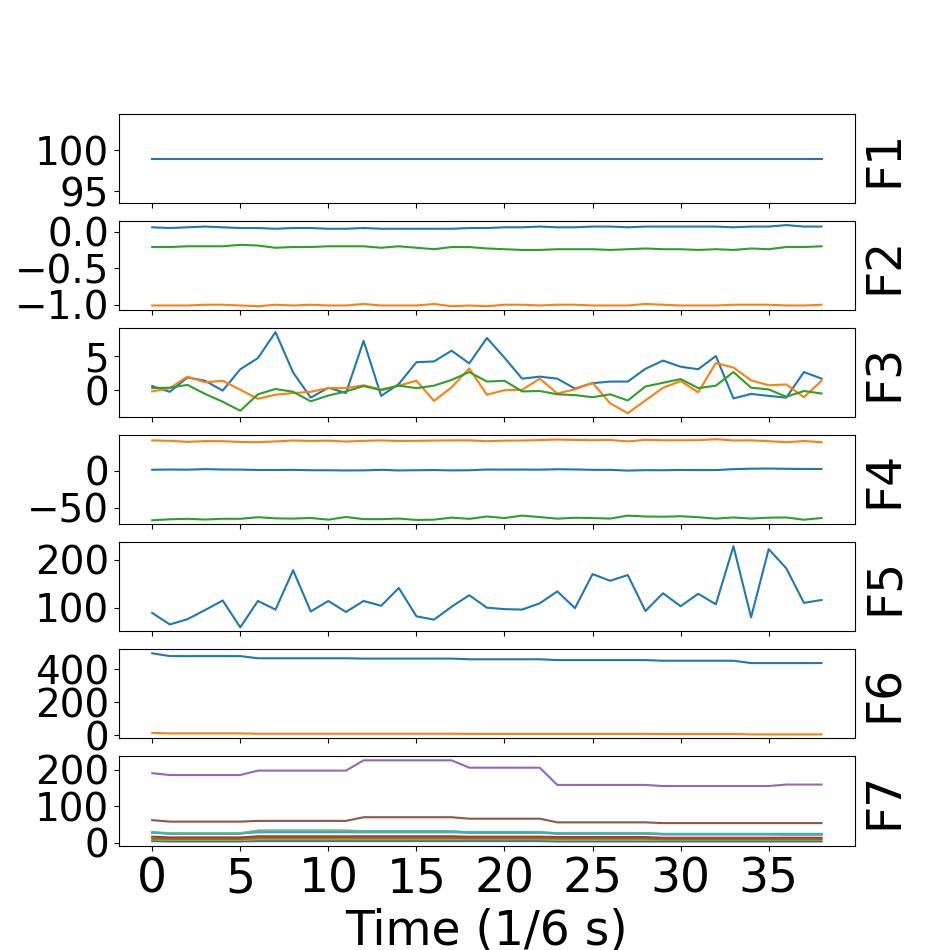}
         \caption{Cutting food}
         \label{fig:label8}
     \end{subfigure}
     \hfill
     \begin{subfigure}[b]{0.24\textwidth}
         \centering
         \includegraphics[width=\textwidth]{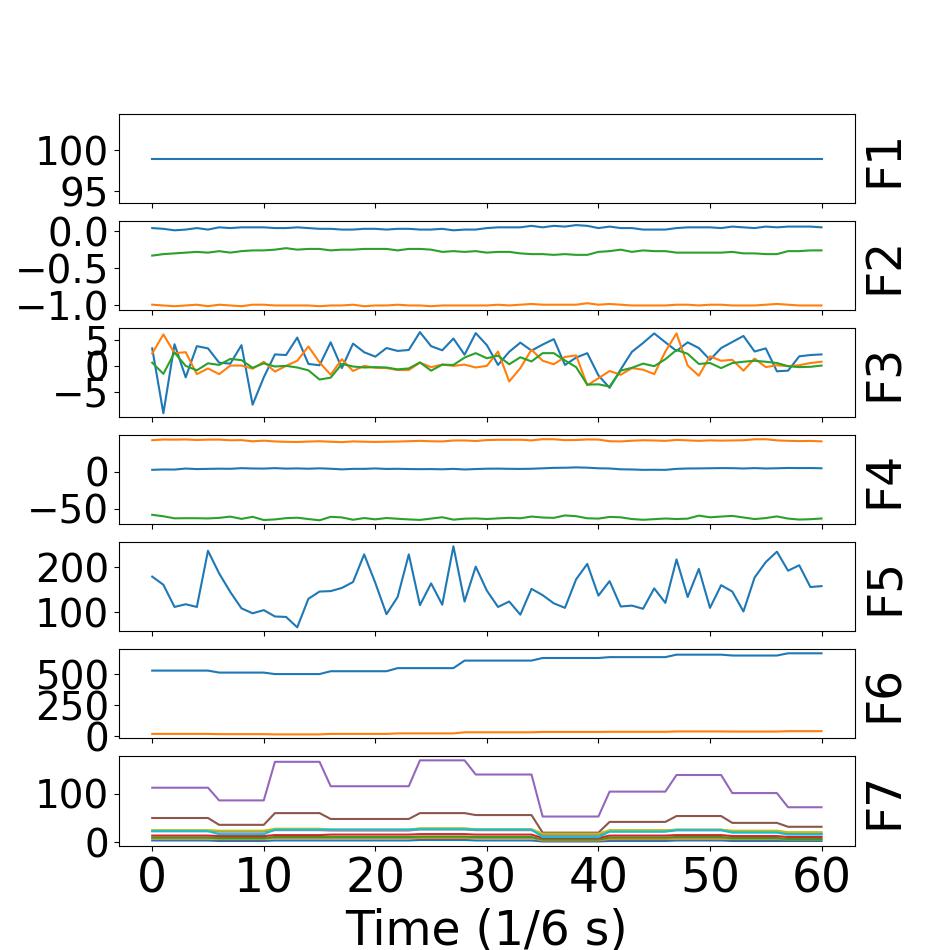}
         \caption{Drinking hot tea}
         \label{fig:label9}
     \end{subfigure}
     \hfill
     \begin{subfigure}[b]{0.24\textwidth}
         \centering
         \includegraphics[width=\textwidth]{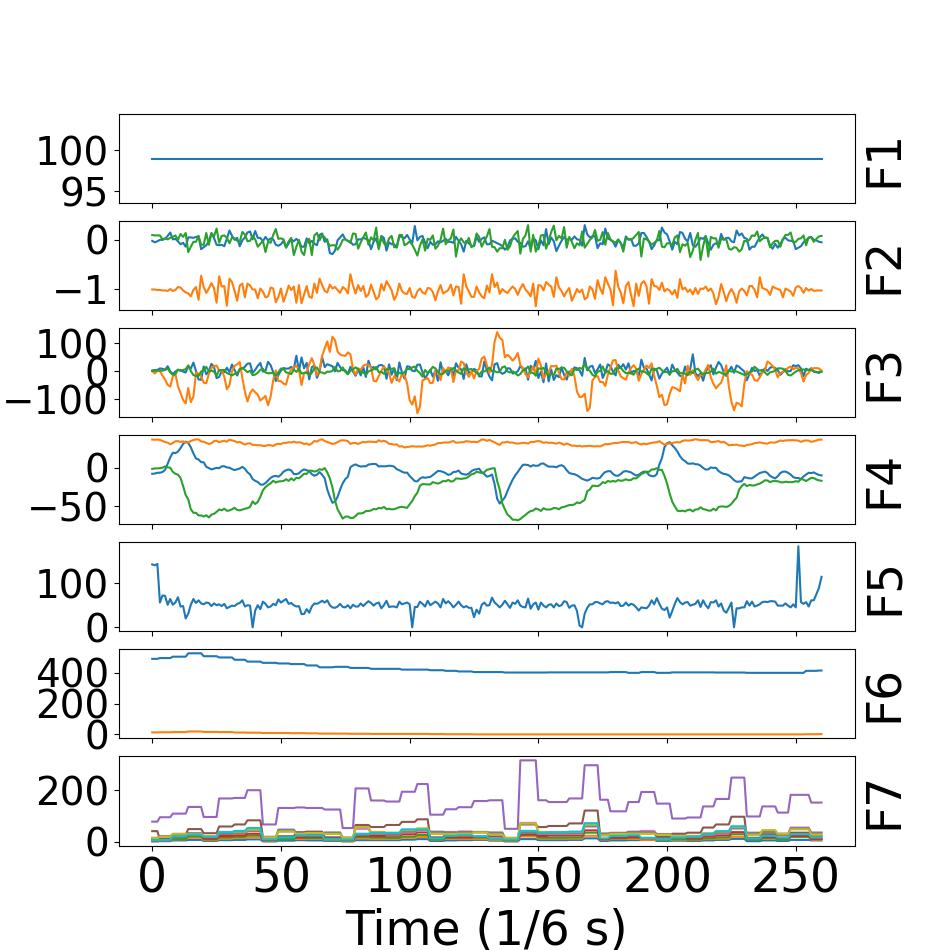}
         \caption{Drinking hot coffee}
         \label{fig:label10}
     \end{subfigure}
     \hfill
     \begin{subfigure}[b]{0.24\textwidth}
         \centering
         \includegraphics[width=\textwidth]{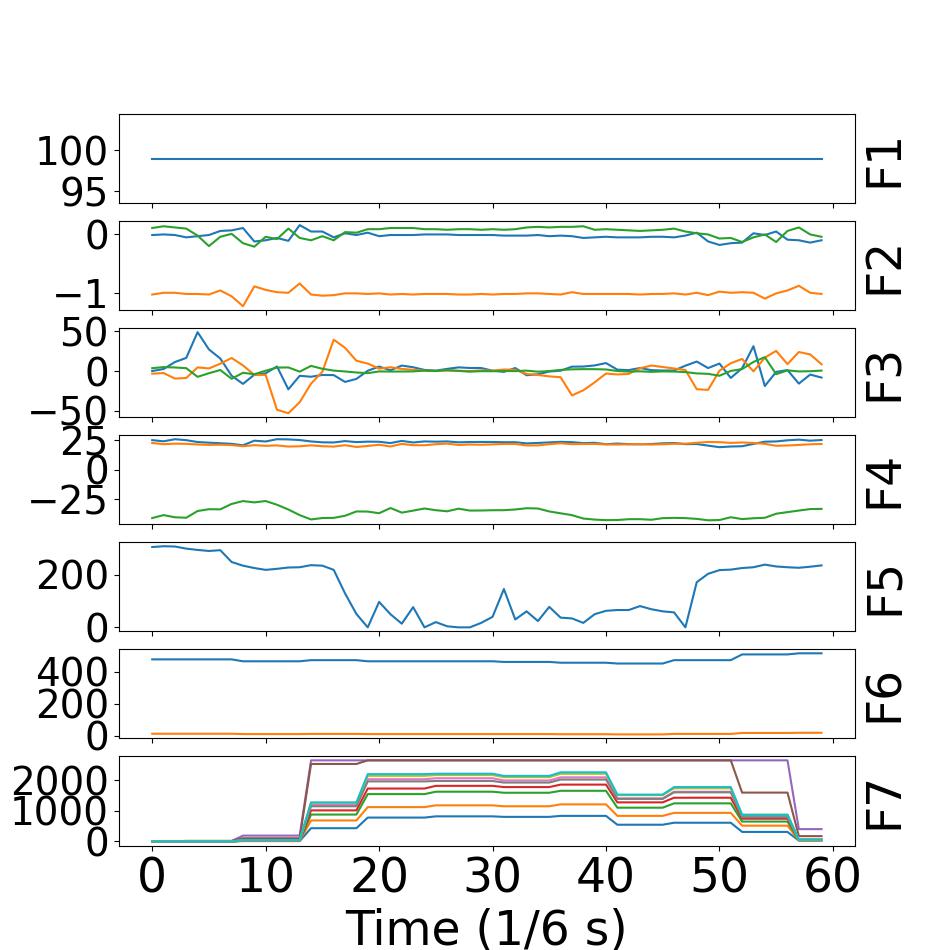}
         \caption{Drinking milk}
         \label{fig:label11}
     \end{subfigure}
     \hfill
     \begin{subfigure}[b]{0.24\textwidth}
         \centering
         \includegraphics[width=\textwidth]{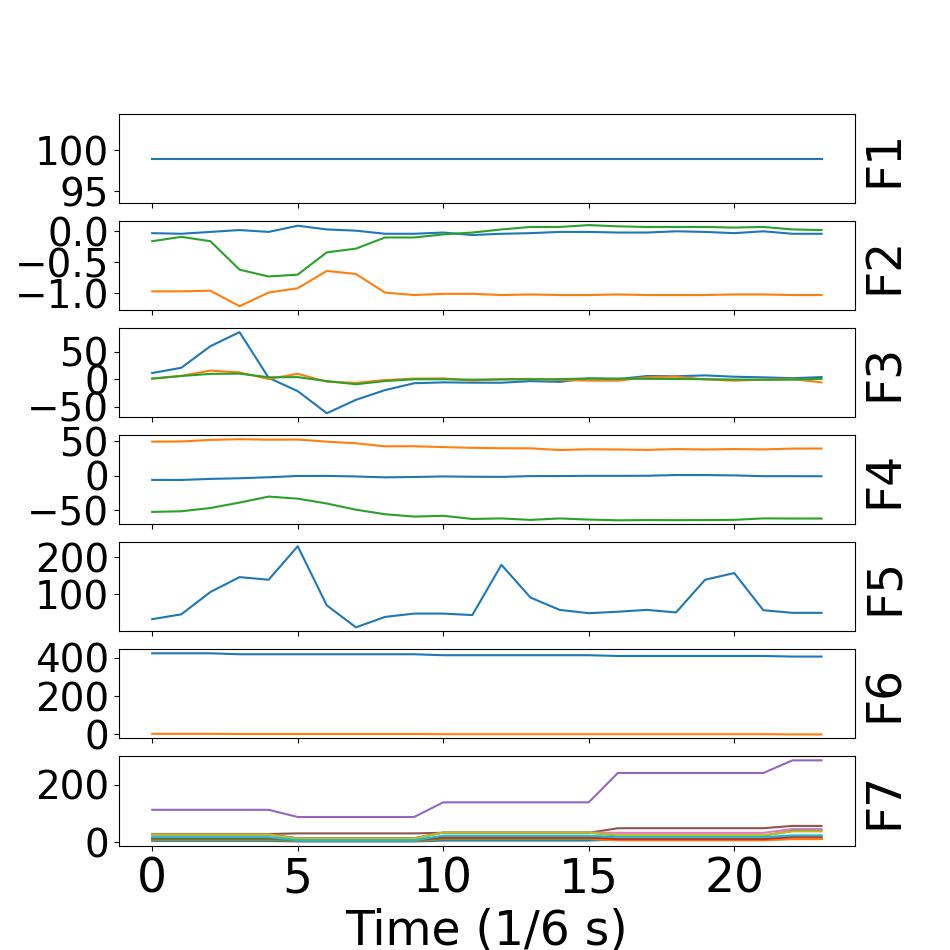}
         \caption{Drinking nature water}
         \label{fig:label12}
     \end{subfigure}
     \hfill
     \begin{subfigure}[b]{0.24\textwidth}
         \centering
         \includegraphics[width=\textwidth]{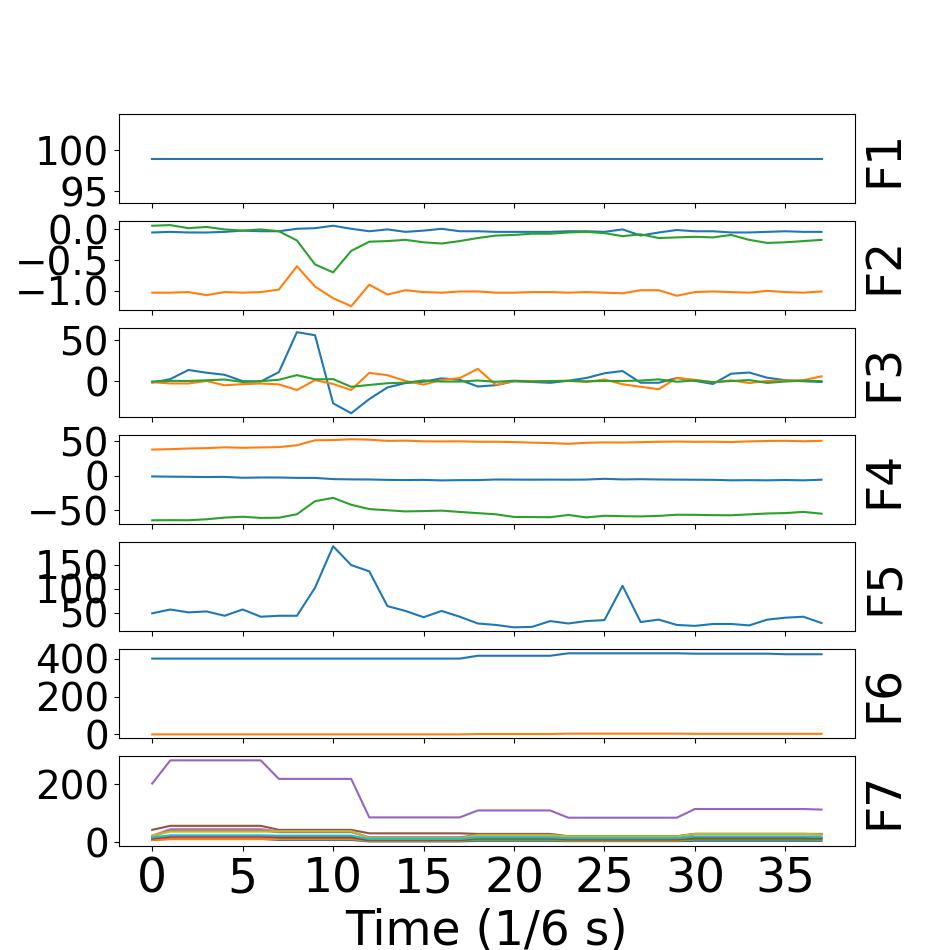}
         \caption{Drinking carbonated water}
         \label{fig:label13}
    \end{subfigure}
    \caption{Raw data when taking 14 activities (\textbf{F1}: pressure, \textbf{F2}: 3 axis Accelerometer data, \textbf{F3}: 3 axis gyrometer data, \textbf{F4}: 3 axis Magnetometer data, \textbf{F5}: distance data, \textbf{F6}: Carbon dioxide (CO2) and Total of Volatile Organic Compounds (TVOC), \textbf{F7}: 10 channel optical data)}
    \label{fig:rawdata}
\end{figure}

\cref{fig:rawdata} presents the raw data of all sensors (except infrared array sensor MLX90640) while the volunteers performed those 14 kinds of activities. In these graphs we can find that single channel sensor data did not present apparent features between different activities. Multiple sensor fusion methods should be applied for those kinds of activity classification. However, optical spectrum sensors provide supplementary information for some drinking-related activities recognition as shown in \cref{fig:label10}, \cref{fig:label11} and \cref{fig:label12}. Since the color between coffee, milk, and water is significantly different and the container of beverages is transparent. When the transparent glass with beverages appears on the chest while volunteer drink, the optical spectrum signals vary for different beverages, which could be very meaningful features for classifier. Although the action of drinking different beverages is very similar, they could still be distinguished through the optical spectrum and movement information.

\begin{table}[hbt]
\renewcommand{\arraystretch}{1.1}
\centering
\footnotesize
\caption{Activity Set}
\label{tab:activity}
\begin{tabular}{|c|l|c|l|}
\hline
Activity ID & Activity & Activity ID & Activity\\
\hline
1& sitting down &8& washing hand \\
\hline
2& standing up &9& cutting food\\
\hline
3& walking &10& drinking hot tea\\
\hline
4& opening microwave oven&11& drinking hot coffee\\
\hline
5& opening freezer&12& drinking milk\\
\hline
6& opening door&13& drinking nature water\\
\hline
7& boiling water&14& drinking carbonated water\\
\hline
\end{tabular}
\end{table}

\subsection{Kitchen activity classification}

\begin{figure}
     \centering
     \begin{subfigure}[b]{0.49\textwidth}
         \centering
         \includegraphics[width=\textwidth]{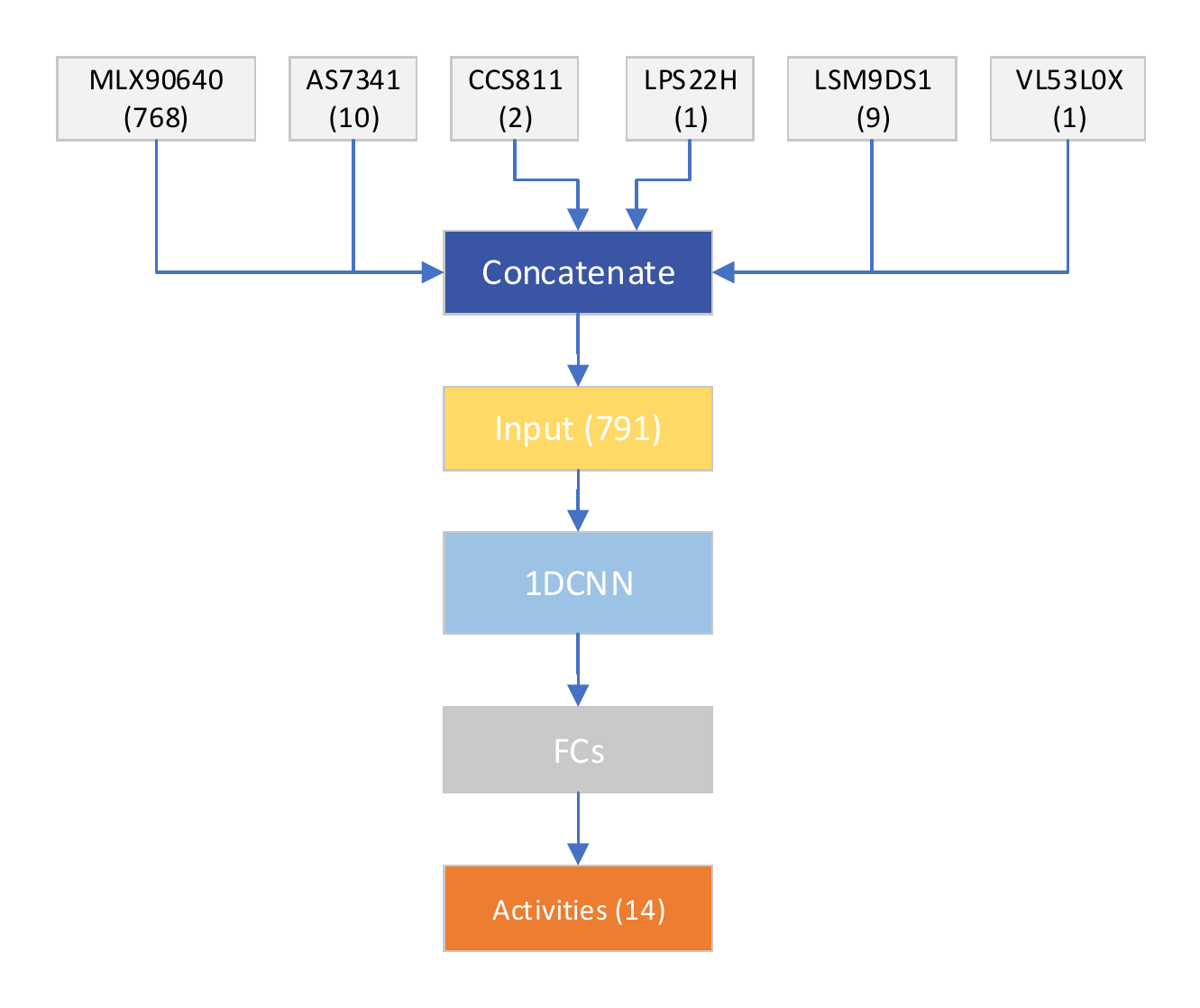}
         \caption{Data fusion}
         \label{fig:dataFusion}
     \end{subfigure}
     \hfill
     \begin{subfigure}[b]{0.49\textwidth}
         \centering
         \includegraphics[width=\textwidth]{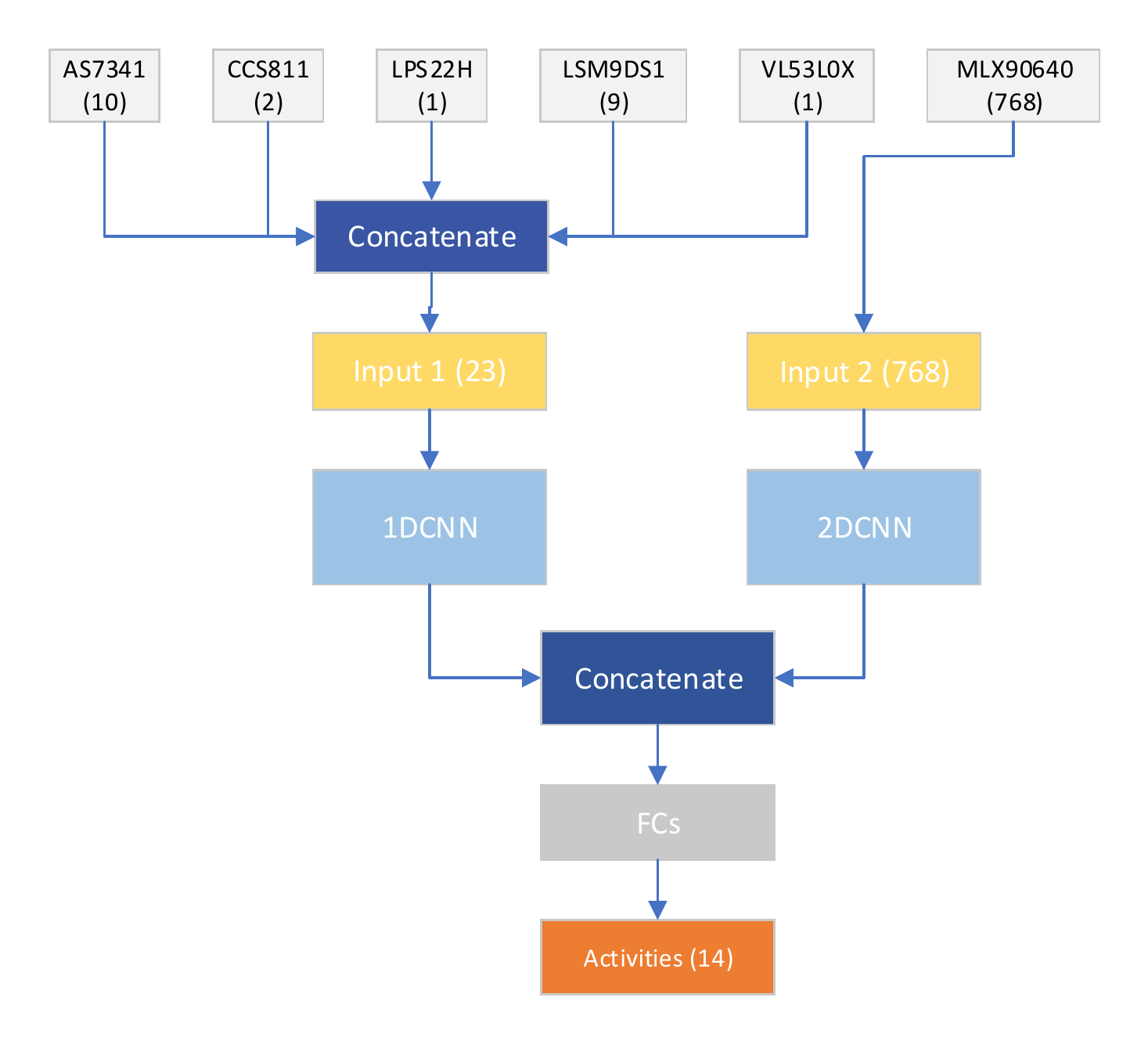}
         \caption{Feature fusion}
         \label{fig:featureFusion}
     \end{subfigure}
        \caption{Architecture of information fusion of sensor data from smart badge}
        \label{fig:informationfusion}
\end{figure}

Since there was data from up to 791 channels obtained from our proposed smart badge, two kinds of information fusion methods, data fusion and feature fusion method based on MC-CNN \cite{yang2015deep} shown in \cref{fig:dataFusion} and \cref{fig:featureFusion}, were applied to classify those 14 activities. For the data fusion method, we concatenated 791 channel data from six sensors directly, which were fed into the 1D CNN composed of three 1D convolutional layers and three max-pooling layers with the window size of 20 and slide step of 10 after normalization to extract features, then the extracted features were input to two fully connect layers to get an activity recognition result. The dataset has 791 channels, but 768 channels are from the infrared array sensor. Thus, it can be easily biased to the features of the infrared array sensor. To prevent the bias and balance between different sensors, we adopt the feature fusion method which has two different feature extraction processes. The first feature extraction network is the same as the data fusion network but is to extract features from all sensors except the infrared array sensor. The second feature extraction is composed of three 2D convolutional layers and max pooling layers and extracts feature only from the infrared array sensor. The infrared array sensor data is image data, so it can be fed into the 2D CNN. The two different features extracted from the 1D CNN and 2D CNN are concatenated and classified by the fully connected layers. The numbers of features from the 1D convolutional neural network and 2D convolutional neural network are not very different. By this feature fusion method, the data from all six different sensors have a similar influence on neural network training at the beginning.

Apart from this, to explore the importance of different sensors for kitchen activities recognition, we divided the original dataset into four sub-datasets according to the number of feature channels, such as 791 channels (including all sensor data), 768 channels (only infrared array sensor data), 23 channels (including all sensor data except infrared array sensor data), 17 channels (including all sensor data except infrared array sensor data, accelerometer and gyroscope data). 

Due to the dataset imbalance, the overall classification accuracy is not an appropriate measure of performance, the F1-score is also utilized to evaluate the kitchen activity classifier, which considers the correct classification of each class equally important. The F1-score is defined by the equation \ref{eqn:f1}, which decided by the precision ($\frac{TP}{TP+FP}$) and the recall ($\frac{TP}{TP+FN}$), where TP, FP are the number of true and false positives, respectively, and FN corresponds to the number of false negatives \cite{ordonez2016deep}. \cref{tab:results} shows a summary of the leave-one-session-out classification result of 14 kinds of human activities. The feature fusion model achieved the highest classification accuracy 92.44 \% with the input of 791 channel data. Although the infrared array sensor has 768 channel data, which is far more than the sum of the other sensor channels combined, the recognition accuracy only using the infrared array sensor is still the lowest. Noteworthy, the 23 channel data without data from the infrared array sensor has also realized a classification accuracy of more than 91 \%, which is very closed to the highest accuracy.  

Overall, in our proposed application scenario, the more sensor modalities the wearable device has, the higher accuracy of activity recognition can be achieved. The feature fusion method is more advantageous than the data fusion method in the activity recognition based on multi-modal sensors. It is worth mentioning that the F1 Score is also very close to accuracy value, though the dataset is severely imbalanced as shown in  \cref{fig:dataset}. 

\begin{equation} \label{eqn:f1}
	F_1score = 2*\frac{precision * recall}{precision + recall}
\end{equation}

\begin{table}[hbt]
\renewcommand{\arraystretch}{1.3}
\centering
\footnotesize
\caption{Results of 14 kinds of human activity recognition}
\label{tab:results}
\begin{tabular}{|c|c|c|c|c|}
\hline
\multirow{2}{*}{Channels} & \multicolumn{2}{|c|}{Data fusion} &\multicolumn{2}{|c|}{Feature fusion}\\
\cline{2-5}
& Accuracy & F1-Score & Accuracy & F1-Score\\
\hline
791 & 90.54 $\pm$ 4.99 & 86.12 $\pm$ 5.85 & \textbf{92.44 $\pm$ 4.33} &\textbf{88.27 $\pm$ 5.27} \\
768 & 82.00 $\pm$ 7.73 & 77.80 $\pm$ 7.80 & 82.88 $\pm$ 6.70 & 77.09 $\pm$7.32 \\
23 & 91.05 $\pm$ 5.27 & 85.95 $\pm$ 6.40 & - & - \\
17 & 89.43 $\pm$ 5.18 & 83.07 $\pm$ 6.50 & - & - \\
\hline
\end{tabular}
\end{table}


\begin{figure}[hbt]
\includegraphics[width=0.7\linewidth, height = 11cm ]{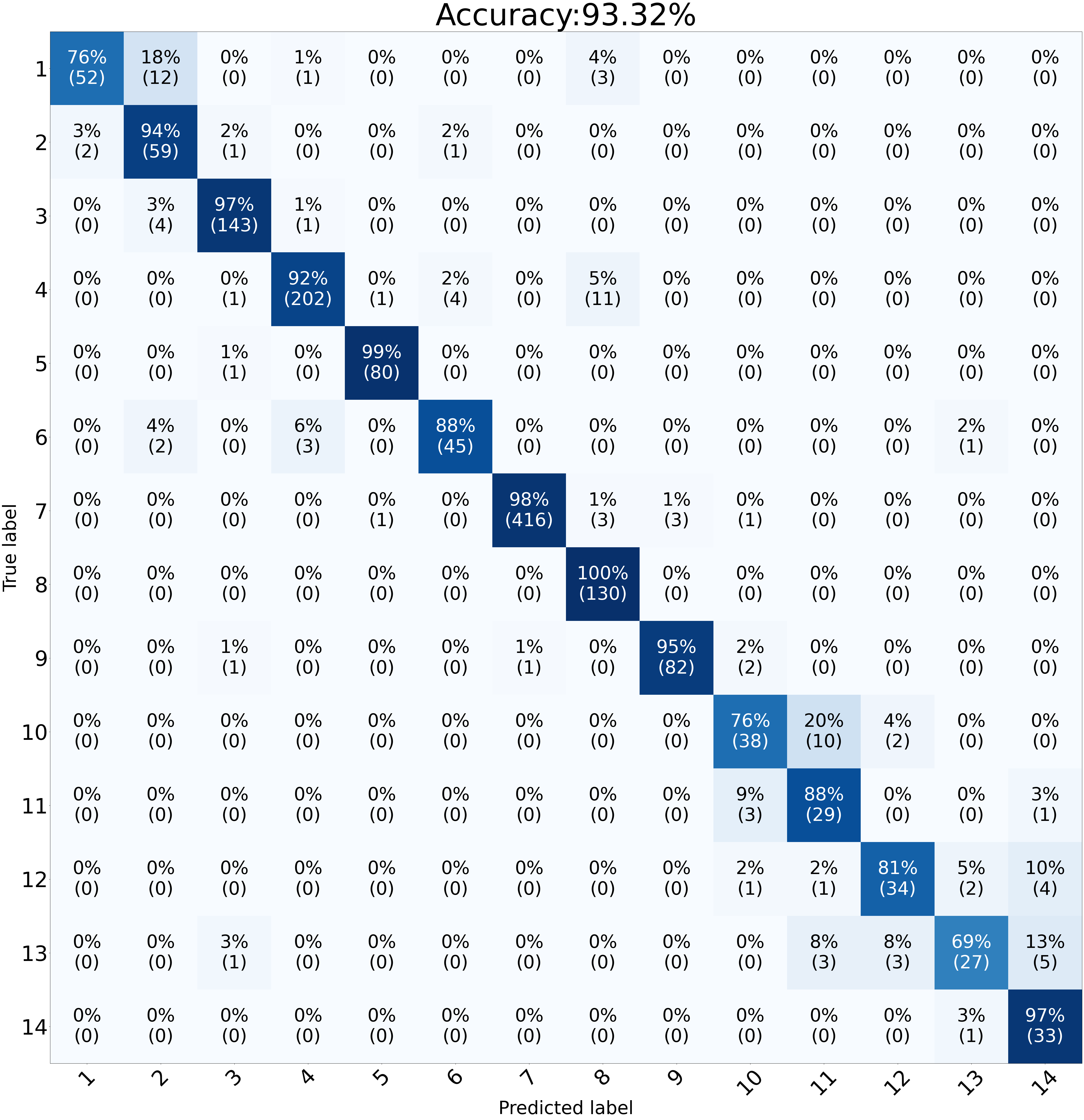}
\captionof{figure}{Confusion matrix from feature fusion MC-CNN of 791 feature channels from 6 sensors}
\label{fig: CM_classfication}
\end{figure}

\section{Discussion}
\label{sec:discussion}

From the confusion matrix of the highest accuracy recognition result shown in \cref{fig: CM_classfication}, it can be observed that the recognition accuracy of drinking-related activities is not as good as other activities classification. Coffee and tea, nature and carbonated water have similar colors. Therefore it is not easy to distinguish them by appearance features from the optical sensor. Although the gas sensor can measure the concentration of carbon dioxide and volatile organic compounds providing useful odor information from the surrounding, nature and carbonated water could be recognized theoretically by the feature of carbon dioxide concentration. The odor features measured by the gas sensor from these two kinds of beverages were still not obvious, as shown in \cref{fig:label12} and \cref{fig:label13}. Because the gas sensor was not very close to these beverages, the concentration of carbon dioxide decreased very fast when carbon dioxide entered the air from the drink. The real carbon dioxide concentration of beverages can not be measured. A new sense modality could be added to improve the recognition accuracy of these kinds of activities in future work. Besides, the recognition result of the sitting down activity was also lower than average accuracy. Because we decreased the sampling rate of all sensors to reduce power consumption, movement-related activities are majorly detected by IMU sensors. The low sampling rate degrades the measurement accuracy of IMU. 
    
\section{Conclusion and Future work}
\label{sec:conclusion}

In this work, we first presented a smart badge with multiple sensors to recognize human activity in a kitchen scenario. Secondly, we applied two information fusion methods to detect 14 activities performed by ten volunteers, which achieved an accuracy of 92.44 \%. We finally compared the performance of different sensors in the 14 activities recognition in the kitchen. Besides, we also recorded a ten-hour long dataset with 791 channel features from six sensors, including ten people performing fourteen kitchen activities. From our experiment results can be found that multiple different sensor modalities are of utmost importance for kitchen activity recognition accuracy by a wearable device in daily life environment as fine-grained activities can not be detected by a single sensor accurately, like drinking coffee, IMU sensor can only detect the drinking activity, but the beverage kinds need to be recognized by other sensors. Multiple sensors provide multiple channel information requiring information fusion methods to obtain the desired results. Our experiment results showed that the feature fusion method achieved a better performance than the data level fusion method in our application scenario based on the smart badge. 

Although this proposed smart badge has demonstrated a competitive performance for kitchen activity recognition, we have also observed that there are still some shortcomings in our approach. Since two microcontrollers were utilized in our smart badge and six sensors were driven. Although the sampling rate has been decreased considerately, the working current was still more than 100 mAh, which reduced the battery life. Besides, the recognition result of several activities was lower than 80 \%, which is caused by the low sampling rate of IMU and the lack of a better sensor modality for beverage kinds recognition. In addition, the proposed hardware platform is not flexible enough to connect more modal sensors due to the limitation of the number of communication interfaces on the microcontrollers.

In the future, we will look into optimizing our hardware platform to overcome the shortcomings mentioned above, then extending the application scenario using this smart badge will be considered, such as manufacturing line and healthcare center. In addition, we will deploy the classification model on a micro-controller to realize real-time processing and protect users’ data from cyber-attack.

\begin{acks}
This work has been supported by the BMBF (German Federal Ministry of Education and Research) in the project Eghi (number 16SV8527). The authors would also like to thank Juan Felipe Vargas Colorado for the contribution of designing the proposed hardware.
\end{acks}

\bibliographystyle{ACM-Reference-Format}
\bibliography{sample-base}


\begin{thebibliography}{31}


\ifx \showCODEN    \undefined \def \showCODEN     #1{\unskip}     \fi
\ifx \showDOI      \undefined \def \showDOI       #1{#1}\fi
\ifx \showISBNx    \undefined \def \showISBNx     #1{\unskip}     \fi
\ifx \showISBNxiii \undefined \def \showISBNxiii  #1{\unskip}     \fi
\ifx \showISSN     \undefined \def \showISSN      #1{\unskip}     \fi
\ifx \showLCCN     \undefined \def \showLCCN      #1{\unskip}     \fi
\ifx \shownote     \undefined \def \shownote      #1{#1}          \fi
\ifx \showarticletitle \undefined \def \showarticletitle #1{#1}   \fi
\ifx \showURL      \undefined \def \showURL       {\relax}        \fi
\providecommand\bibfield[2]{#2}
\providecommand\bibinfo[2]{#2}
\providecommand\natexlab[1]{#1}
\providecommand\showeprint[2][]{arXiv:#2}

\bibitem[Aguileta et~al\mbox{.}(2019)]%
        {aguileta2019multi}
\bibfield{author}{\bibinfo{person}{Antonio~A Aguileta},
  \bibinfo{person}{Ramon~F Brena}, \bibinfo{person}{Oscar Mayora},
  \bibinfo{person}{Erik Molino-Minero-Re}, {and} \bibinfo{person}{Luis~A
  Trejo}.} \bibinfo{year}{2019}\natexlab{}.
\newblock \showarticletitle{Multi-sensor fusion for activity recognition—A
  survey}.
\newblock \bibinfo{journal}{\emph{Sensors}} \bibinfo{volume}{19},
  \bibinfo{number}{17} (\bibinfo{year}{2019}), \bibinfo{pages}{3808}.
\newblock


\bibitem[Bansal et~al\mbox{.}(2013)]%
        {bansal2013kitchen}
\bibfield{author}{\bibinfo{person}{Shubham Bansal}, \bibinfo{person}{Shubham
  Khandelwal}, \bibinfo{person}{Shubham Gupta}, {and} \bibinfo{person}{Dushyant
  Goyal}.} \bibinfo{year}{2013}\natexlab{}.
\newblock \showarticletitle{Kitchen activity recognition based on scene
  context}. In \bibinfo{booktitle}{\emph{2013 IEEE International Conference on
  Image Processing}}. \bibinfo{publisher}{IEEE}, \bibinfo{address}{Melbourne,
  VIC, Australia}, \bibinfo{pages}{3461--3465}.
\newblock


\bibitem[Bazzano et~al\mbox{.}(2002)]%
        {bazzano2002fruit}
\bibfield{author}{\bibinfo{person}{Lydia~A Bazzano}, \bibinfo{person}{Jiang
  He}, \bibinfo{person}{Lorraine~G Ogden}, \bibinfo{person}{Catherine~M Loria},
  \bibinfo{person}{Suma Vupputuri}, \bibinfo{person}{Leann Myers}, {and}
  \bibinfo{person}{Paul~K Whelton}.} \bibinfo{year}{2002}\natexlab{}.
\newblock \showarticletitle{Fruit and vegetable intake and risk of
  cardiovascular disease in US adults: the first National Health and Nutrition
  Examination Survey Epidemiologic Follow-up Study}.
\newblock \bibinfo{journal}{\emph{The American journal of clinical nutrition}}
  \bibinfo{volume}{76}, \bibinfo{number}{1} (\bibinfo{year}{2002}),
  \bibinfo{pages}{93--99}.
\newblock


\bibitem[Bharti et~al\mbox{.}(2018)]%
        {bharti2018human}
\bibfield{author}{\bibinfo{person}{Pratool Bharti}, \bibinfo{person}{Debraj
  De}, \bibinfo{person}{Sriram Chellappan}, {and} \bibinfo{person}{Sajal~K
  Das}.} \bibinfo{year}{2018}\natexlab{}.
\newblock \showarticletitle{HuMAn: Complex activity recognition with
  multi-modal multi-positional body sensing}.
\newblock \bibinfo{journal}{\emph{IEEE Transactions on Mobile Computing}}
  \bibinfo{volume}{18}, \bibinfo{number}{4} (\bibinfo{year}{2018}),
  \bibinfo{pages}{857--870}.
\newblock


\bibitem[Bian et~al\mbox{.}(2022)]%
        {bian2022state}
\bibfield{author}{\bibinfo{person}{Sizhen Bian}, \bibinfo{person}{Mengxi Liu},
  \bibinfo{person}{Bo Zhou}, {and} \bibinfo{person}{Paul Lukowicz}.}
  \bibinfo{year}{2022}\natexlab{}.
\newblock \showarticletitle{The State-of-the-Art Sensing Techniques in Human
  Activity Recognition: A Survey}.
\newblock \bibinfo{journal}{\emph{Sensors}} \bibinfo{volume}{22},
  \bibinfo{number}{12} (\bibinfo{year}{2022}), \bibinfo{pages}{4596}.
\newblock


\bibitem[Bianchi et~al\mbox{.}(2019)]%
        {bianchi2019iot}
\bibfield{author}{\bibinfo{person}{Valentina Bianchi}, \bibinfo{person}{Marco
  Bassoli}, \bibinfo{person}{Gianfranco Lombardo}, \bibinfo{person}{Paolo
  Fornacciari}, \bibinfo{person}{Monica Mordonini}, {and}
  \bibinfo{person}{Ilaria De~Munari}.} \bibinfo{year}{2019}\natexlab{}.
\newblock \showarticletitle{IoT wearable sensor and deep learning: An
  integrated approach for personalized human activity recognition in a smart
  home environment}.
\newblock \bibinfo{journal}{\emph{IEEE Internet of Things Journal}}
  \bibinfo{volume}{6}, \bibinfo{number}{5} (\bibinfo{year}{2019}),
  \bibinfo{pages}{8553--8562}.
\newblock


\bibitem[Cheng et~al\mbox{.}(2010)]%
        {cheng2010active}
\bibfield{author}{\bibinfo{person}{Jingyuan Cheng}, \bibinfo{person}{Oliver
  Amft}, {and} \bibinfo{person}{Paul Lukowicz}.}
  \bibinfo{year}{2010}\natexlab{}.
\newblock \showarticletitle{Active capacitive sensing: Exploring a new wearable
  sensing modality for activity recognition}. In
  \bibinfo{booktitle}{\emph{International conference on pervasive computing}}.
  \bibinfo{publisher}{Springer}, \bibinfo{address}{Berlin, Heidelberg},
  \bibinfo{pages}{319--336}.
\newblock


\bibitem[Du et~al\mbox{.}(2015)]%
        {du2015hierarchical}
\bibfield{author}{\bibinfo{person}{Yong Du}, \bibinfo{person}{Wei Wang}, {and}
  \bibinfo{person}{Liang Wang}.} \bibinfo{year}{2015}\natexlab{}.
\newblock \showarticletitle{Hierarchical recurrent neural network for skeleton
  based action recognition}. In \bibinfo{booktitle}{\emph{Proceedings of the
  IEEE conference on computer vision and pattern recognition}}.
  \bibinfo{publisher}{IEEE}, \bibinfo{address}{Boston, MA, USA},
  \bibinfo{pages}{1110--1118}.
\newblock


\bibitem[Gravina et~al\mbox{.}(2017)]%
        {gravina2017multi}
\bibfield{author}{\bibinfo{person}{Raffaele Gravina}, \bibinfo{person}{Parastoo
  Alinia}, \bibinfo{person}{Hassan Ghasemzadeh}, {and}
  \bibinfo{person}{Giancarlo Fortino}.} \bibinfo{year}{2017}\natexlab{}.
\newblock \showarticletitle{Multi-sensor fusion in body sensor networks:
  State-of-the-art and research challenges}.
\newblock \bibinfo{journal}{\emph{Information Fusion}}  \bibinfo{volume}{35}
  (\bibinfo{year}{2017}), \bibinfo{pages}{68--80}.
\newblock


\bibitem[Gravina and Li(2019)]%
        {gravina2019emotion}
\bibfield{author}{\bibinfo{person}{Raffaele Gravina} {and}
  \bibinfo{person}{Qimeng Li}.} \bibinfo{year}{2019}\natexlab{}.
\newblock \showarticletitle{Emotion-relevant activity recognition based on
  smart cushion using multi-sensor fusion}.
\newblock \bibinfo{journal}{\emph{Information Fusion}}  \bibinfo{volume}{48}
  (\bibinfo{year}{2019}), \bibinfo{pages}{1--10}.
\newblock


\bibitem[G{\"u}nthermann et~al\mbox{.}(2022)]%
        {gunthermann2022slow}
\bibfield{author}{\bibinfo{person}{Lukas G{\"u}nthermann},
  \bibinfo{person}{Lloyd Pellatt}, {and} \bibinfo{person}{Daniel Roggen}.}
  \bibinfo{year}{2022}\natexlab{}.
\newblock \showarticletitle{Slow Feature Preprocessing in Deep Neural Networks
  for Wearable Sensor-Based Locomotion Recognition}. In
  \bibinfo{booktitle}{\emph{2022 IEEE International Conference on Pervasive
  Computing and Communications Workshops and other Affiliated Events (PerCom
  Workshops)}}. \bibinfo{publisher}{IEEE}, \bibinfo{address}{Pisa, Italy},
  \bibinfo{pages}{58--61}.
\newblock


\bibitem[Jiang et~al\mbox{.}(2020)]%
        {jiang2020novel}
\bibfield{author}{\bibinfo{person}{Shuo Jiang}, \bibinfo{person}{Qinghua Gao},
  \bibinfo{person}{Huaiyang Liu}, {and} \bibinfo{person}{Peter~B Shull}.}
  \bibinfo{year}{2020}\natexlab{}.
\newblock \showarticletitle{A novel, co-located EMG-FMG-sensing wearable
  armband for hand gesture recognition}.
\newblock \bibinfo{journal}{\emph{Sensors and Actuators A: Physical}}
  \bibinfo{volume}{301} (\bibinfo{year}{2020}), \bibinfo{pages}{111738}.
\newblock


\bibitem[Kamachi et~al\mbox{.}(2021)]%
        {kamachi2021prediction}
\bibfield{author}{\bibinfo{person}{Haruka Kamachi}, \bibinfo{person}{Tahera
  Hossain}, \bibinfo{person}{Fuyuka Tokuyama}, \bibinfo{person}{Anna Yokokubo},
  {and} \bibinfo{person}{Guillaume Lopez}.} \bibinfo{year}{2021}\natexlab{}.
\newblock \showarticletitle{Prediction of eating activity using smartwatch}. In
  \bibinfo{booktitle}{\emph{Adjunct Proceedings of the 2021 ACM International
  Joint Conference on Pervasive and Ubiquitous Computing and Proceedings of the
  2021 ACM International Symposium on Wearable Computers}}.
  \bibinfo{publisher}{ACM}, \bibinfo{address}{Virtual},
  \bibinfo{pages}{304--309}.
\newblock


\bibitem[Kehler et~al\mbox{.}(2018)]%
        {kehler2018systematic}
\bibfield{author}{\bibinfo{person}{D~Scott Kehler},
  \bibinfo{person}{Jacqueline~L Hay}, \bibinfo{person}{Andrew~N Stammers},
  \bibinfo{person}{Naomi~C Hamm}, \bibinfo{person}{Dustin~E Kimber},
  \bibinfo{person}{Annette~SH Schultz}, \bibinfo{person}{Andrea Szwajcer},
  \bibinfo{person}{Rakesh~C Arora}, \bibinfo{person}{Navdeep Tangri}, {and}
  \bibinfo{person}{Todd~A Duhamel}.} \bibinfo{year}{2018}\natexlab{}.
\newblock \showarticletitle{A systematic review of the association between
  sedentary behaviors with frailty}.
\newblock \bibinfo{journal}{\emph{Experimental gerontology}}
  \bibinfo{volume}{114} (\bibinfo{year}{2018}), \bibinfo{pages}{1--12}.
\newblock


\bibitem[Lei et~al\mbox{.}(2012)]%
        {lei2012fine}
\bibfield{author}{\bibinfo{person}{Jinna Lei}, \bibinfo{person}{Xiaofeng Ren},
  {and} \bibinfo{person}{Dieter Fox}.} \bibinfo{year}{2012}\natexlab{}.
\newblock \showarticletitle{Fine-grained kitchen activity recognition using
  rgb-d}. In \bibinfo{booktitle}{\emph{Proceedings of the 2012 ACM Conference
  on Ubiquitous Computing}}. \bibinfo{publisher}{ACM},
  \bibinfo{address}{Pittsburgh, USA}, \bibinfo{pages}{208--211}.
\newblock


\bibitem[Luo et~al\mbox{.}(2019)]%
        {luo2019kitchen}
\bibfield{author}{\bibinfo{person}{Fei Luo}, \bibinfo{person}{Stefan Poslad},
  {and} \bibinfo{person}{Eliane Bodanese}.} \bibinfo{year}{2019}\natexlab{}.
\newblock \showarticletitle{Kitchen activity detection for healthcare using a
  low-power radar-enabled sensor network}. In \bibinfo{booktitle}{\emph{ICC
  2019-2019 IEEE International Conference on Communications (ICC)}}.
  \bibinfo{publisher}{IEEE}, \bibinfo{pages}{1--7}.
\newblock


\bibitem[Matsuyama et~al\mbox{.}(2021)]%
        {matsuyama2021deep}
\bibfield{author}{\bibinfo{person}{Hitoshi Matsuyama},
  \bibinfo{person}{Shunsuke Aoki}, \bibinfo{person}{Takuro Yonezawa},
  \bibinfo{person}{Kei Hiroi}, \bibinfo{person}{Katsuhiko Kaji}, {and}
  \bibinfo{person}{Nobuo Kawaguchi}.} \bibinfo{year}{2021}\natexlab{}.
\newblock \showarticletitle{Deep Learning for Ballroom Dance Recognition: A
  Temporal and Trajectory-Aware Classification Model With Three-Dimensional
  Pose Estimation and Wearable Sensing}.
\newblock \bibinfo{journal}{\emph{IEEE Sensors Journal}} \bibinfo{volume}{21},
  \bibinfo{number}{22} (\bibinfo{year}{2021}), \bibinfo{pages}{25437--25448}.
\newblock


\bibitem[Mayer and Thomas(1967)]%
        {mayer1967regulation}
\bibfield{author}{\bibinfo{person}{Jean Mayer} {and} \bibinfo{person}{Donald~W
  Thomas}.} \bibinfo{year}{1967}\natexlab{}.
\newblock \showarticletitle{Regulation of Food Intake and Obesity: The
  regulation of food intake is complex; a number of abnormalities may cause
  obesity.}
\newblock \bibinfo{journal}{\emph{Science}} \bibinfo{volume}{156},
  \bibinfo{number}{3773} (\bibinfo{year}{1967}), \bibinfo{pages}{328--337}.
\newblock


\bibitem[Mohammad et~al\mbox{.}(2017)]%
        {mohammad2017dataset}
\bibfield{author}{\bibinfo{person}{Yasser Mohammad}, \bibinfo{person}{Kazunori
  Matsumoto}, {and} \bibinfo{person}{Keiichiro Hoashi}.}
  \bibinfo{year}{2017}\natexlab{}.
\newblock \showarticletitle{A dataset for activity recognition in an unmodified
  kitchen using smart-watch accelerometers}. In
  \bibinfo{booktitle}{\emph{Proceedings of the 16th International Conference on
  Mobile and Ubiquitous Multimedia}}. \bibinfo{publisher}{ACM},
  \bibinfo{address}{Stuttgart, Germany}, \bibinfo{pages}{63--68}.
\newblock


\bibitem[Natarajan and Nevatia(2008)]%
        {natarajan2008online}
\bibfield{author}{\bibinfo{person}{Pradeep Natarajan} {and}
  \bibinfo{person}{Ramakant Nevatia}.} \bibinfo{year}{2008}\natexlab{}.
\newblock \showarticletitle{Online, real-time tracking and recognition of human
  actions}. In \bibinfo{booktitle}{\emph{2008 IEEE Workshop on Motion and video
  Computing}}. \bibinfo{publisher}{IEEE}, \bibinfo{pages}{1--8}.
\newblock


\bibitem[Oguntala et~al\mbox{.}(2019)]%
        {oguntala2019smartwall}
\bibfield{author}{\bibinfo{person}{George~A Oguntala}, \bibinfo{person}{Raed~A
  Abd-Alhameed}, \bibinfo{person}{Nazar~T Ali}, \bibinfo{person}{Yim-Fun Hu},
  \bibinfo{person}{James~M Noras}, \bibinfo{person}{Nnabuike~N Eya},
  \bibinfo{person}{Issa Elfergani}, {and} \bibinfo{person}{Jonathan
  Rodriguez}.} \bibinfo{year}{2019}\natexlab{}.
\newblock \showarticletitle{SmartWall: Novel RFID-enabled ambient human
  activity recognition using machine learning for unobtrusive health
  monitoring}.
\newblock \bibinfo{journal}{\emph{IEEE Access}}  \bibinfo{volume}{7}
  (\bibinfo{year}{2019}), \bibinfo{pages}{68022--68033}.
\newblock


\bibitem[Ord{\'o}{\~n}ez and Roggen(2016)]%
        {ordonez2016deep}
\bibfield{author}{\bibinfo{person}{Francisco~Javier Ord{\'o}{\~n}ez} {and}
  \bibinfo{person}{Daniel Roggen}.} \bibinfo{year}{2016}\natexlab{}.
\newblock \showarticletitle{Deep convolutional and lstm recurrent neural
  networks for multimodal wearable activity recognition}.
\newblock \bibinfo{journal}{\emph{Sensors}} \bibinfo{volume}{16},
  \bibinfo{number}{1} (\bibinfo{year}{2016}), \bibinfo{pages}{115}.
\newblock


\bibitem[Qiu et~al\mbox{.}(2022)]%
        {qiu2022multi}
\bibfield{author}{\bibinfo{person}{Sen Qiu}, \bibinfo{person}{Hongkai Zhao},
  \bibinfo{person}{Nan Jiang}, \bibinfo{person}{Zhelong Wang},
  \bibinfo{person}{Long Liu}, \bibinfo{person}{Yi An}, \bibinfo{person}{Hongyu
  Zhao}, \bibinfo{person}{Xin Miao}, \bibinfo{person}{Ruichen Liu}, {and}
  \bibinfo{person}{Giancarlo Fortino}.} \bibinfo{year}{2022}\natexlab{}.
\newblock \showarticletitle{Multi-sensor information fusion based on machine
  learning for real applications in human activity recognition:
  State-of-the-art and research challenges}.
\newblock \bibinfo{journal}{\emph{Information Fusion}}  \bibinfo{volume}{80}
  (\bibinfo{year}{2022}), \bibinfo{pages}{241--265}.
\newblock


\bibitem[Volkow et~al\mbox{.}(2011)]%
        {volkow2011reward}
\bibfield{author}{\bibinfo{person}{Nora~D Volkow}, \bibinfo{person}{Gene-Jack
  Wang}, {and} \bibinfo{person}{Ruben~D Baler}.}
  \bibinfo{year}{2011}\natexlab{}.
\newblock \showarticletitle{Reward, dopamine and the control of food intake:
  implications for obesity}.
\newblock \bibinfo{journal}{\emph{Trends in cognitive sciences}}
  \bibinfo{volume}{15}, \bibinfo{number}{1} (\bibinfo{year}{2011}),
  \bibinfo{pages}{37--46}.
\newblock


\bibitem[Wang et~al\mbox{.}(2019)]%
        {wang2019deep}
\bibfield{author}{\bibinfo{person}{Jindong Wang}, \bibinfo{person}{Yiqiang
  Chen}, \bibinfo{person}{Shuji Hao}, \bibinfo{person}{Xiaohui Peng}, {and}
  \bibinfo{person}{Lisha Hu}.} \bibinfo{year}{2019}\natexlab{}.
\newblock \showarticletitle{Deep learning for sensor-based activity
  recognition: A survey}.
\newblock \bibinfo{journal}{\emph{Pattern recognition letters}}
  \bibinfo{volume}{119} (\bibinfo{year}{2019}), \bibinfo{pages}{3--11}.
\newblock


\bibitem[Xu et~al\mbox{.}(2019)]%
        {xu2019innohar}
\bibfield{author}{\bibinfo{person}{Cheng Xu}, \bibinfo{person}{Duo Chai},
  \bibinfo{person}{Jie He}, \bibinfo{person}{Xiaotong Zhang}, {and}
  \bibinfo{person}{Shihong Duan}.} \bibinfo{year}{2019}\natexlab{}.
\newblock \showarticletitle{InnoHAR: A deep neural network for complex human
  activity recognition}.
\newblock \bibinfo{journal}{\emph{Ieee Access}}  \bibinfo{volume}{7}
  (\bibinfo{year}{2019}), \bibinfo{pages}{9893--9902}.
\newblock


\bibitem[Yang et~al\mbox{.}(2015)]%
        {yang2015deep}
\bibfield{author}{\bibinfo{person}{Jianbo Yang}, \bibinfo{person}{Minh~Nhut
  Nguyen}, \bibinfo{person}{Phyo~Phyo San}, \bibinfo{person}{Xiao~Li Li}, {and}
  \bibinfo{person}{Shonali Krishnaswamy}.} \bibinfo{year}{2015}\natexlab{}.
\newblock \showarticletitle{Deep convolutional neural networks on multichannel
  time series for human activity recognition}. In
  \bibinfo{booktitle}{\emph{Twenty-fourth international joint conference on
  artificial intelligence}}. \bibinfo{publisher}{ACM}, \bibinfo{address}{Buenos
  Aires, Argentina}, \bibinfo{pages}{3995–4001}.
\newblock


\bibitem[Yang and Tian(2016)]%
        {yang2016super}
\bibfield{author}{\bibinfo{person}{Xiaodong Yang} {and} \bibinfo{person}{YingLi
  Tian}.} \bibinfo{year}{2016}\natexlab{}.
\newblock \showarticletitle{Super normal vector for human activity recognition
  with depth cameras}.
\newblock \bibinfo{journal}{\emph{IEEE transactions on pattern analysis and
  machine intelligence}} \bibinfo{volume}{39}, \bibinfo{number}{5}
  (\bibinfo{year}{2016}), \bibinfo{pages}{1028--1039}.
\newblock


\bibitem[Yang et~al\mbox{.}(2012)]%
        {yang2012recognizing}
\bibfield{author}{\bibinfo{person}{Xiaodong Yang}, \bibinfo{person}{Chenyang
  Zhang}, {and} \bibinfo{person}{YingLi Tian}.}
  \bibinfo{year}{2012}\natexlab{}.
\newblock \showarticletitle{Recognizing actions using depth motion maps-based
  histograms of oriented gradients}. In \bibinfo{booktitle}{\emph{Proceedings
  of the 20th ACM international conference on Multimedia}}.
  \bibinfo{publisher}{ACM}, \bibinfo{address}{Nara, Japan},
  \bibinfo{pages}{1057--1060}.
\newblock


\bibitem[Zhang et~al\mbox{.}(2020)]%
        {zhang2020necksense}
\bibfield{author}{\bibinfo{person}{Shibo Zhang}, \bibinfo{person}{Yuqi Zhao},
  \bibinfo{person}{Dzung~Tri Nguyen}, \bibinfo{person}{Runsheng Xu},
  \bibinfo{person}{Sougata Sen}, \bibinfo{person}{Josiah Hester}, {and}
  \bibinfo{person}{Nabil Alshurafa}.} \bibinfo{year}{2020}\natexlab{}.
\newblock \showarticletitle{Necksense: A multi-sensor necklace for detecting
  eating activities in free-living conditions}.
\newblock \bibinfo{journal}{\emph{Proceedings of the ACM on interactive,
  mobile, wearable and ubiquitous technologies}} \bibinfo{volume}{4},
  \bibinfo{number}{2} (\bibinfo{year}{2020}), \bibinfo{pages}{1--26}.
\newblock


\bibitem[Zhu et~al\mbox{.}(2018)]%
        {zhu2018indoor}
\bibfield{author}{\bibinfo{person}{Shangyue Zhu}, \bibinfo{person}{Junhong Xu},
  \bibinfo{person}{Hanqing Guo}, \bibinfo{person}{Qiwei Liu},
  \bibinfo{person}{Shaoen Wu}, {and} \bibinfo{person}{Honggang Wang}.}
  \bibinfo{year}{2018}\natexlab{}.
\newblock \showarticletitle{Indoor human activity recognition based on ambient
  radar with signal processing and machine learning}. In
  \bibinfo{booktitle}{\emph{2018 IEEE international conference on
  communications (ICC)}}. \bibinfo{publisher}{IEEE}, \bibinfo{address}{Kansas
  City, MO, USA}, \bibinfo{pages}{1--6}.
\newblock


\end{thebibliography}

\end{document}